\documentclass{article}
\usepackage{iclr2027_conference,times}

\usepackage{amsmath,amsfonts,bm}

\def\eqref#1{equation~\ref{#1}}

\def\1{\bm{1}}

\DeclareMathAlphabet{\mathsfit}{\encodingdefault}{\sfdefault}{m}{sl}
\SetMathAlphabet{\mathsfit}{bold}{\encodingdefault}{\sfdefault}{bx}{n}

\usepackage{amssymb}
\usepackage{float}
\usepackage{colortbl}
\usepackage{graphicx}
\usepackage{url}
\usepackage{hyperref}
\hypersetup{hidelinks}
\newcommand{\best}[1]{\cellcolor{blue!15}\textbf{#1}}

\title{VABench: Measuring Embodied Spatial
Intelligence through Visual Demonstrations,
Active Perception, and Metric Control}

\author{
\textbf{Zhongbo Zhang$^{1,*}$, Jiayi Jin$^{1,*}$, Yifan Wang$^1$, Zaibin Zhang$^1$, Haiwen Diao$^2$}\\
\textbf{Lijun Wang$^1$, Huchuan Lu$^1$}\\[1pt]
{\normalfont $^1$Dalian University of Technology\quad $^2$Nanyang Technological University\quad $^*$Equal contribution.}
}

\makeatletter
\renewcommand{\@maketitle}{%
  \vbox{\hsize\textwidth
    \setlength{\parskip}{0pt}%
    \raggedright
    {\normalfont\bfseries\fontsize{17}{18.5}\selectfont\@title\par}%
    \vskip 8pt
    {\normalfont\fontsize{10}{12}\selectfont\@author\par}%
  }%
}
\makeatother

\iclrfinalcopy
\begin{document}

\maketitle
\lhead{} 
\begingroup
\setlength{\intextsep}{5pt}
\begin{figure}[H]
    \centering
    \includegraphics[width=\linewidth]{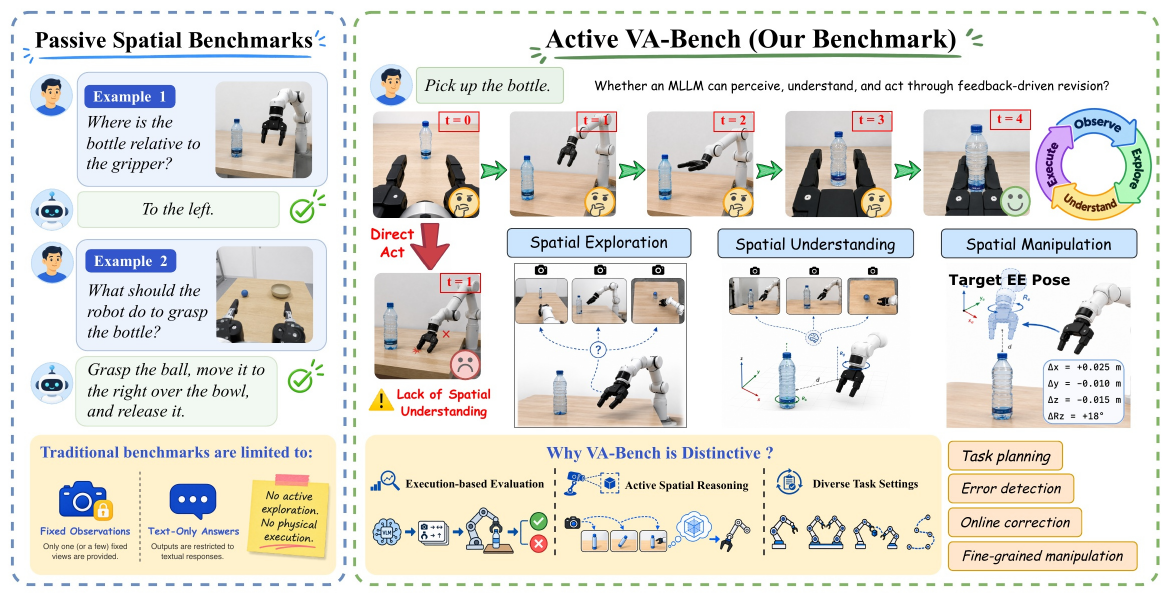}
    \caption{Overview of the evaluation target. Fixed-view spatial evaluations
    can stop at a verbal description, whereas VA-Bench requires an MLLM to
    actively acquire evidence, infer robot--object geometry, issue metric
    manipulation commands, and revise them from execution outcomes.}
    \label{fig:overview}
\end{figure}
\endgroup

\begin{abstract}
Spatial intelligence requires more than describing object locations.
Under incomplete observation, models must identify and acquire missing evidence,
interpret it in a common spatial frame, and act on it. We introduce
\textsc{VA-Bench} to evaluate the complete observe--reason--act--revise loop.
General-purpose MLLMs learn procedural context from RGB-only demonstrations,
actively select camera viewpoints, issue metric Cartesian commands, and revise
them from execution feedback. Models receive no privileged object poses, oracle
trajectories, or learned action heads. A fixed model-agnostic controller executes
only model-specified targets. VA-Bench contains 14 base task
families (11 single-arm and three dual-arm), seven held-out geometry/layout
variants, and a long-horizon five-object composition track. We evaluate 12
primary model conditions in three independent runs over the same 20 physically
verified seeds per base task, reporting terminal success, nine trajectory-level
behavioral diagnostics, and subtask progress.
\textbf{First, the best-performing model scores 100.0\% on target localization
and 78.9\% on spatial relations in the annotated run. Its three-run
macro-average task success is only
$53.93\!\pm\!3.17\%$.}
\textbf{Second, active camera control significantly improves task success over
passive multi-view observation. In one matched comparison, success rises
from 27.86\% to 57.50\%.}
\textbf{Third, held-out geometric transfer can reduce task success by over
30 percentage points. No model completes a strict long-horizon episode,
despite substantial partial progress.} VA-Bench thus tests whether
general-purpose MLLMs can turn visual demonstrations and actively acquired
evidence into successful embodied action. Code and benchmark:
\href{https://github.com/zhangzhongbo2213/VABench}{\nolinkurl{github.com/zhangzhongbo2213/VABench}}.
\end{abstract}

\section{Introduction}
\label{sec:introduction}

Spatial intelligence is not limited to describing where objects appear in an
image. Under partial observability, an agent must determine what information is
missing, acquire it, and interpret it in a common spatial frame before acting
\citep{liu2023visualspatialreasoning,kamath2023whatsup,du2024embspatialbench,song2025robospatial}.
This requirement is especially important for manipulation: viewpoint, distance,
and occlusion can make the metric relationship between the robot and an object
ambiguous in a single view, and pixel displacement alone does not determine
motion in world coordinates
\citep{fu2024blink,yang2025thinkinginspace,ma2025_3dsrbench,azuma2022scanqa,ma2023sqa3d}.
Reliable interaction therefore depends on active
evidence acquisition and contact-level spatial estimation
\citep{chen2024spatialvlm,cheng2024spatialrgpt,liao2024quantitativespatial,nasiriany2024pivot,yuan2024robopoint,huang2024rekep}.

Recent benchmarks extend spatial evaluation from fixed- and multi-view
reasoning to embodied planning
\citep{fu2024blink,yang2025thinkinginspace,song2025robospatial,zheng2022vlmbench,
zhao2025manipbench,zhang2025vlabench,zhang2026theoryspace,hong2026esibench,
gao2026spatialworld,yang2025embodiedbench,wu2026stbibench,maurya2026imbench}.
They test important components of spatial intelligence. Evidence acquisition,
inference, and execution, however, remain coupled. Each observation or action
changes the evidence for the next decision
\citep{huang2023innermonologue,nasiriany2024pivot,huang2024rekep,
zhang2026theoryspace,hong2026esibench}. This motivates our central question:
\emph{when a model must close this loop on its own, can spatial understanding
support reliable, executable action under the constraints of physical
interaction?}

Figure~\ref{fig:overview} illustrates the evaluation gap that motivates
VA-Bench. A model may describe a spatial relation correctly without estimating
geometry precisely enough for contact
\citep{chen2024spatialvlm,liao2024quantitativespatial,nasiriany2024pivot,yuan2024robopoint},
while preselected views do not reveal
whether it can identify and resolve missing evidence
\citep{fu2024blink,yang2025thinkinginspace,ma2025_3dsrbench,
zhang2026theoryspace,hong2026esibench}. Likewise, predefined
skills or policies may execute a task without testing whether the model can
convert spatial judgments into metric robot motion
\citep{ichter2023saycan,huang2023innermonologue,
liang2023codeaspolicies,driess2023palme,rana2023sayplan,
huang2023voxposer,nasiriany2024pivot,fang2024moka,
huang2024rekep}. VA-Bench evaluates these
capabilities jointly within a single closed-loop interaction.

We introduce \textsc{VA-Bench}, a physics-simulated benchmark for this complete
observe--reason--act--revise loop. Given RGB observations and task
instructions, general-purpose MLLMs select viewpoints and produce metric
Cartesian commands. They receive no privileged object poses, oracle
trajectories, or task waypoints. A fixed model-agnostic controller only executes
the targets specified by the MLLM. A successful RGB demonstration supplies
procedural context but no geometry for the current test instance. The evaluation
therefore includes both procedure interpretation and instance-specific
re-grounding.

VA-Bench contains 14 task families, including 11 single-arm and three dual-arm
families. They span grasping, placement, tool use, handover, and coordinated
manipulation. Each primary condition is evaluated in three independent runs over the same 20
deterministic, physically verified seeds per task, yielding 840 run--seed
episodes scored by terminal task success or failure. We further test controlled
geometry and layout transfer and a five-object long-horizon task. Nine
trajectory-level behavioral diagnostics and subtask-progress measures localize observable degradation along
the interaction loop without attributing every failure to spatial reasoning
alone.

Our evaluation of 12 primary model conditions yields three main findings.
\textbf{First, strong target localization does not ensure reliable task
execution.} The best-performing model attains $100.0\%$ on target localization
and $78.9\%$ on spatial relations in the annotated run. Its three-run 14-task
macro-average success is only $53.93\!\pm\!3.17\%$.
\textbf{Second, active camera control improves task success over passive
multi-view observation.} In a matched comparison, success rises from
$27.86\%$ to $57.50\%$ with active camera control. Across all 12 conditions,
the best active-exploration diagnostic score is $83.2\%$.
\textbf{Third, held-out geometric transfer and long-horizon composition remain
challenging.} Across seven matched families, transfer degradation varies
substantially: one condition drops $32.14$ points, from $80.00\%$ to $47.86\%$,
while another drops $10.00$ points, from $77.86\%$ to $67.86\%$. The strongest
conditions complete 61/100 and 53/100 object placements, respectively, but
neither completes a single strict long-horizon episode.

Our contributions are threefold:
\begin{itemize}
    \item We introduce \textsc{VA-Bench}, which evaluates whether a
    general-purpose MLLM can interpret an RGB demonstration, actively acquire
    evidence, ground it into metric single- or dual-arm actions, and revise
    those actions from execution outcomes.

    \item We construct 14 task families with controlled held-out variants and a
    long-horizon compositional task, together with nine behavioral diagnostics
    and subtask-progress measures for localizing failures within the closed loop.

    \item We evaluate 12 primary model conditions over three runs and
    characterize three limitations of current closed-loop competence: the
    mismatch between spatial diagnostics and execution, degradation under
    geometric transfer, and failure to compose atomic progress into
    long-horizon completion.
\end{itemize}


\section{Related Work}
\label{sec:related-work}

\subsection{Spatial Reasoning under Partial Observation}
Spatial reasoning has been studied extensively using static images, supplied multi-view observations, and 3D scene representations. CLEVR and GQA established benchmarks for compositional visual reasoning~\citep{johnson2017clevr,hudson2019gqa}. SpatialVLM develops qualitative and quantitative spatial understanding~\citep{chen2024spatialvlm}, while CV-Bench, introduced with Cambrian-1, and subsequent evaluations probe spatial relations, depth, distance, and scene understanding in multimodal models~\citep{tong2024cambrian,fu2024blink,yang2025thinkinginspace,ma2025_3dsrbench}. ManipBench brings this diagnosis closer to manipulation by testing low-level reasoning from supplied observations, without an interactive execution loop~\citep{zhao2025manipbench}.

Theory of Space, ESI-Bench, and SpatialWorld make observation gathering interactive, evaluating spatial beliefs or task decisions as agents explore partially observed environments~\citep{zhang2026theoryspace,hong2026esibench,gao2026spatialworld}. Their evaluations focus primarily on spatial beliefs,
semantic decisions, or high-level task outcomes, rather than directly grounding
acquired evidence into metric end-effector control with execution-level feedback.
VA-Bench bridges this gap by requiring the same model to acquire task-relevant
visual evidence, translate spatial understanding into metric arm motions, and
revise subsequent decisions from execution outcomes.

\subsection{Embodied Manipulation and Robot Benchmarks}
Meta-World, RLBench, CALVIN, LIBERO, and ManiSkill provide standardized environments for evaluating manipulation learning and generalization~\citep{yu2020metaworld,james2020rlbench,mees2022calvin,liu2023libero,mu2021maniskill,gu2023maniskill2}; BEHAVIOR-1K and RoboCasa extend evaluation to diverse household activities~\citep{li2022behavior1k,nasiriany2024robocasa}. A complementary policy-learning line, including RT-1, RT-2, OpenVLA, and Octo, learns robot action generation from demonstration data~\citep{brohan2023rt1,zitkovich2023rt2,kim2024openvla,ghosh2024octo}. These results establish strong visuomotor control capabilities, but their evaluation generally assumes learned policy execution or specialized action-generation modules rather than requiring a general-purpose MLLM to close the observation-to-action loop.

More directly related benchmarks already test general-purpose MLLMs in executed manipulation. EmbodiedBench's manipulation setting supplies a fixed front view with object annotations and 3D coordinates~\citep{yang2025embodiedbench}. ST-BiBench includes continuous bimanual control but relies on auxiliary
ground-truth pose information~\citep{wu2026stbibench}. IMBench’s action stage evaluates both a vision-based agent and an
object-state-augmented agent through parameterized motor primitives rather than
direct metric action generation~\citep{maurya2026imbench}. ESPIRE evaluates localization and manipulation execution through generated spatial outputs, with depth-based lifting from image coordinates to robot poses~\citep{zhao2026espire}. These benchmarks provide important progress toward executable manipulation
evaluation. However, they typically rely on structured spatial information,
object states, or predefined execution components. VA-Bench instead evaluates
whether the same general-purpose MLLM can actively select observations and
ground them into metric arm commands within a joint view--arm loop, without
supplied object geometry or a learned robot action head.

\subsection{Active perception and demonstration-grounded control}
Active perception and next-best-view planning study how sensing actions reduce uncertainty or improve geometric coverage~\citep{bajcsy1988active,pito1999nextbestview}. Their integration with manipulation also has direct precedents: Vision in Action learns task-relevant camera behavior and bimanual control from human demonstrations~\citep{xiong2025visioninaction}. Such work establishes the value of joint sensing and control through learned visuomotor policies; VA-Bench examines this coordination through explicit decisions made by a general-purpose MLLM.

Demonstration-based learning likewise provides established routes from observed behavior to action. RoboMimic studies learning from recorded robot demonstrations, BC-Z uses language or video task conditioning, and MimicPlay combines human play videos with robot demonstrations~\citep{mandlekar2021whatmatters,jang2021bcz,wang2023mimicplay}. MimicGen scales training data by adapting demonstration segments to new scenes~\citep{mandlekar2023mimicgen}. These systems use action-bearing robot data to train control policies. In contrast, VA-Bench treats demonstrations as contextual guidance rather than training signals, requiring the evaluated model to infer reusable procedures from visual
demonstrations and re-ground them in unseen scenes. In VA-Bench, the evaluated model first extracts a textual procedure from sampled RGB demonstration frames; execution must instantiate that procedure in the current scene without demonstration trajectories, object poses, or task-specific policy training.

Together, these lines motivate the setting summarized in Table~\ref{tab:related-work}: a \emph{joint view--arm loop for general-purpose MLLMs without learned robot action heads}. VA-Bench studies whether general-purpose MLLMs can integrate demonstration-derived procedures, active observation, and metric manipulation within a unified closed-loop protocol.

\begin{table}[htbp]
\centering
\caption{Comparison of representative evaluation settings. Entries describe the cited protocols, not every capability of their underlying simulators.}
\label{tab:related-work}
\begingroup
\footnotesize
\setlength{\tabcolsep}{4pt}
\renewcommand{\arraystretch}{1.13}
\begin{tabular*}{\linewidth}{@{}l@{\extracolsep{\fill}}cccc@{}}
\hline
Setting & \shortstack{Active\\observation} & \shortstack{Executed\\arm control} & \shortstack{Joint\\view--arm loop} & \shortstack{General MLLM,\\no learned head} \\
\hline
ManipBench                  & No  & No  & No  & Yes \\
Theory of Space             & Yes & No  & No  & Yes \\
ESI-Bench                   & Yes & No  & No  & Yes \\
SpatialWorld                & Yes & No  & No  & Yes \\
EmbodiedBench (manipulation)& No  & Yes & No  & Yes \\
ST-BiBench (fine control)   & No  & Yes & No  & Yes \\
IMBench (action stage)      & No  & Yes & No  & Yes \\
ESPIRE (execution)          & No  & Yes & No  & Yes \\
\textbf{VA-Bench}           & Yes & Yes & Yes & Yes \\
\hline
\end{tabular*}
\par\vspace{3pt}
\begin{minipage}{\linewidth}
\scriptsize
\emph{Active observation} means explicit view-acquisition decisions, including navigation; incidental camera motion is excluded.
\emph{Executed arm control} includes parameterized motor primitives and geometrically lifted predictions, but excludes spatial QA and symbolic object interactions.
A \emph{joint view--arm loop} interleaves viewpoint and arm-control decisions within manipulation.
The last column permits fixed controllers and does not imply identical observation inputs.
\end{minipage}
\endgroup
\end{table}

In contrast to prior benchmarks that evaluate spatial understanding,
manipulation execution, or active perception separately, VA-Bench studies their
interaction through a joint view--arm loop within a single closed-loop protocol,
where an MLLM must acquire missing evidence, ground spatial concepts into metric
actions, and adapt behavior based on physical feedback.

\section{VA-Bench: Benchmark Design and Evaluation Protocol}
\label{sec:benchmark}

\textsc{VA-Bench}, the \emph{Vision-Action Benchmark}, evaluates
general-purpose MLLMs as embodied agents in physics-based simulation.  In the
standard protocol, a model infers task context from visual demonstrations,
acquires visual evidence, generates metric robot commands, and revises them
from execution feedback.  Each evaluation episode is scored by a physics-based
task-success predicate.

\subsection{Evaluation tracks and task suite}
\label{sec:benchmark-tracks}

\paragraph{Base track.}
The base suite contains 14 task families, including 11 single-arm and 3
dual-arm families, with 20 physically verified scene seeds per family.
Table~\ref{tab:task-coverage} groups them by their primary spatial and
manipulation demands.  Each family receives equal weight in the base score.

\begin{table}[t]
\caption{Coverage of the 14 VA-Bench base task families.  Counts are the
number of families in each analysis group.}
\label{tab:task-coverage}
\centering
\begingroup
\small
\setlength{\tabcolsep}{3pt}
\begin{tabular}{@{}p{1.35in}p{0.35in}p{3.50in}@{}}
\hline
\textbf{Task group} & \textbf{No.} & \textbf{Primary spatial and manipulation demand} \\
\hline
Direct grasp grounding & 5 & Locate a graspable region, approach it with the correct end-effector orientation, and close at a usable height/depth. \\
Placement and stacking & 4 & Track object--container or object--support relations and control release height, orientation, and final contact. \\
Functional contact and tool use & 2 & Select a contact point and approach direction for pressing, striking, or other goal-directed interaction. \\
Bimanual coordination & 3 & Assign arm roles, maintain relative geometry, and sequence synchronized or handover actions. \\
\hline
\end{tabular}
\endgroup
\end{table}

\paragraph{Controlled transfer track.}
Seven single-arm families have held-out variants that change object geometry,
the container or box instance, or the layout while preserving the instruction.
They test re-grounding of a learned procedure in a changed scene.

\paragraph{Long-horizon composition track.}
The five-object basket task combines four atomic manipulation skills into a
longer sequence of placements that requires sustained progress tracking.  Its
demonstration structure and horizon differ from the base suite, so we report
it separately.

All three tracks allow model-directed viewpoint selection.  The passive-view
control removes this choice and is reported as a separate ablation
(Section~\ref{sec:camera-protocol-ablation}).

\subsection{Interactive episode protocol}
\label{sec:benchmark-formulation}

Each episode specifies a task family $\tau$, a fixed scene seed, a task
instruction, demonstration context, an action budget $B_{\tau}$, and a
success predicate $\phi_{\tau}$.  The seed determines the scene but provides
no privileged geometry to the model.  Appendix~\ref{sec:app-task-formulation}
gives the full instance specification.

Let $x_t$ denote the physical simulator state.  At decision step $t$, the
observation consists of the current active-camera image $I_t$, filtered robot
proprioception $p_t$, interaction history $h_t$, and the previous
action/planner outcome $f_{t-1}$.  The robot and camera action sets are
$\mathcal{A}_{\mathrm{robot}}$ and $\mathcal{A}_{\mathrm{cam}}$:
\begin{equation}
    o_t=(I_t,p_t,h_t,f_{t-1}), \qquad
    \begin{cases}
    x_{t+1}=T_{\tau}(x_t,a_t), & a_t\in\mathcal{A}_{\mathrm{robot}},\\
    x_{t+1}=x_t, & a_t\in\mathcal{A}_{\mathrm{cam}}.
    \end{cases}
    \label{eq:benchmark-loop}
\end{equation}
Robot actions change the physical state; camera actions change only the
observation pose.  The model may also call a non-environment tool or stop,
neither of which advances the environment.

Camera and robot actions consume the same episode budget.  Indexing dispatched
camera and robot actions by $t$, the episode ends at
$t_{\mathrm{end}}\leq B_{\tau}$ when
$\phi_{\tau}$ is satisfied, the model stops, or the budget is exhausted.

\subsection{Demonstration-conditioned re-grounding}
\label{sec:demonstration-protocol}

In the standard base protocol, model $m$ inspects multiple distinct frames
from an RGB-only video of a successful task execution and writes a structured
textual summary $s_{m,\tau}$.  It infers constraints such as grasp region,
approach, and arm coordination from the visible procedure.  Neither the video
nor the summary supplies machine-readable actions, robot or object poses,
contact labels, semantic phases, or a metric solution for the live scene.
The model must re-ground the procedure under a new seed and generate every
numerical action parameter
\citep{jang2021bcz,jiang2023vima,mandlekar2023mimicgen,
wang2023mimicplay,mandlekar2021whatmatters}.

Held-out transfer variants reuse the base-task summary without a new video.
The long-horizon basket task instead provides four atomic-skill videos and no
full-task demonstration.  Demonstrations may be omitted only in explicitly
named controls.  Appendix~\ref{sec:app-demonstration-protocol} gives the
summary protocol and controls, and Appendix~\ref{app:agent-harness-details}
describes the frame-inspection gate.

\subsection{Observation, action, and privilege boundary}
\label{sec:benchmark-action-interface}

\paragraph{Model-visible inputs.}
The model receives the task instruction, demonstration-derived summary, clean
RGB, and filtered proprioception: end-effector pose, gripper state, finger
midpoint (GC), and gripper-local axes in world coordinates, for each controlled
arm.  History retains prior actions and textual feedback, including action
validity and trajectory-planning outcomes.

\paragraph{Robot and camera actions.}
$\mathcal{A}_{\mathrm{robot}}$ is parameterized.  The model supplies the
magnitude of each world-axis translation (1--100~mm) and gripper-local rotation
(1--90~degrees), and issues gripper aperture commands.  Dual-arm tasks also
support synchronized variants.  These numerical targets are generated by the
model, not selected from a predefined menu.
$\mathcal{A}_{\mathrm{cam}}$ is a bounded discrete set of gripper-centered
semantic viewpoints and fixed-step local translation, zoom, yaw, and pitch
adjustments.  Camera actions test \emph{where} the model chooses to look;
robot actions require metric spatial estimation.  The camera interface does
not control a physical camera arm.
Appendices~\ref{sec:app-robot-action-space}--\ref{sec:app-camera-action-space}
give the complete schemas and bounds.

\paragraph{Withheld information.}
The model does not receive target markers, keypoints, masks, depth,
segmentation, object coordinates or poses, contact or task-checker state, task
waypoints, camera calibration, offline seed-verification probes, or seed
provenance.  These signals remain evaluator-only.

\subsection{Agent harness and fixed execution}
\label{sec:agent-harness}

All models share a harness that maintains interaction history, validates
decisions, and dispatches commands.  The MLLM is the only model-dependent
component, with no learned robot policy or model-specific action head.
The fixed executor checks reachability and executes trajectories.  The model
selects objects, grasp regions, arm roles, waypoints, and corrective actions.
The primary direct-control condition
uses no learned RGB-D or pre-grasp spatial module.  Any such module is reported
as a named ablation with its own policy and input manifest.
Appendix~\ref{app:agent-harness-details} gives the implementation details and
illustrates the interaction loop in Figure~\ref{fig:agent-harness}.

\subsection{Task construction and instance validation}
\label{sec:benchmark-construction}

The suite is implemented in RoboTwin with a fixed, model-agnostic inverse
kinematics and trajectory layer \citep{mu2025robotwin}.  Candidate task families must have a visually
recoverable spatial bottleneck and support direct control, a deterministic
success checker, and observable subtask checkpoints.

Seed construction is deterministic.  Each family retains 20 seeds after
settling and physical-verification checks.  All 280 retained instances pass
a complete expert execution from the robot home state to terminal success.
The suite is fixed before evaluation and reused unchanged across models and
runs.  Verification states and actions remain evaluator-only.
Appendix~\ref{sec:app-seed-validation} gives the screening protocol, probed
counts, randomization ranges, and terminal predicates.

\subsection{Outcomes and behavioral diagnostics}
\label{sec:diagnostic-scoring}

For model $m$, run $r$, task $\tau$, and valid seed $i$, terminal success is
the binary outcome of the task-specific RoboTwin checker:
\begin{equation}
    y_{m,r,\tau,i}=\mathbf{1}\!\left[\phi_{\tau}(x_{t_{\mathrm{end}}})=1\right],
    \qquad t_{\mathrm{end}}\leq B_{\tau}.
    \label{eq:terminal-outcome}
\end{equation}
RoboTwin checks the predicate after each dispatched action.  Stops and budget
exhaustion without success count as failures, with no post-hoc exclusions.
The outcome measures the complete model--interface--executor system.

We complement terminal success with nine trajectory-level behavioral criteria,
human-annotated on one complete base-suite run per evaluated condition and
grouped into three modules.  \emph{Spatial perception and understanding} includes
target perception and localization (TL), informative active exploration (AE),
and robot--object spatial relations (SR).  \emph{Robot manipulation} includes
manipulation-semantics understanding (MS), goal-directed manipulation planning
(MP), and fine-grained pre-contact analysis (FG).  \emph{Error recovery}
includes error detection (ED), online correction (OC), and post-failure
adjustment (PF).  These criteria help locate failures in the trajectory and
complement the environment checker
\citep{liu2023reflect,ren2023robotsaskhelp,duan2025aha,
xiong2024aicmllm,jiang2024transic}.

Each episode label is the majority vote of three annotators.  ED, OC, and PF
apply only when an observable error occurs; inapplicable episodes are NA and
excluded from their denominators.  The other six criteria apply to every
annotated base episode.  Terminal success remains the primary outcome, and
module means are descriptive equal-weight summaries.
Appendix~\ref{sec:app-diagnostic-scoring} gives the rubric, subtask-progress
measure, and annotation examples.  Agreement and vote-rule sensitivity appear
in Appendix~\ref{sec:app-annotation-reliability}.

\section{Main Results and Analysis}
\label{sec:main-results}

\begin{figure}[!htbp]
    \centering
    \includegraphics[width=\linewidth]{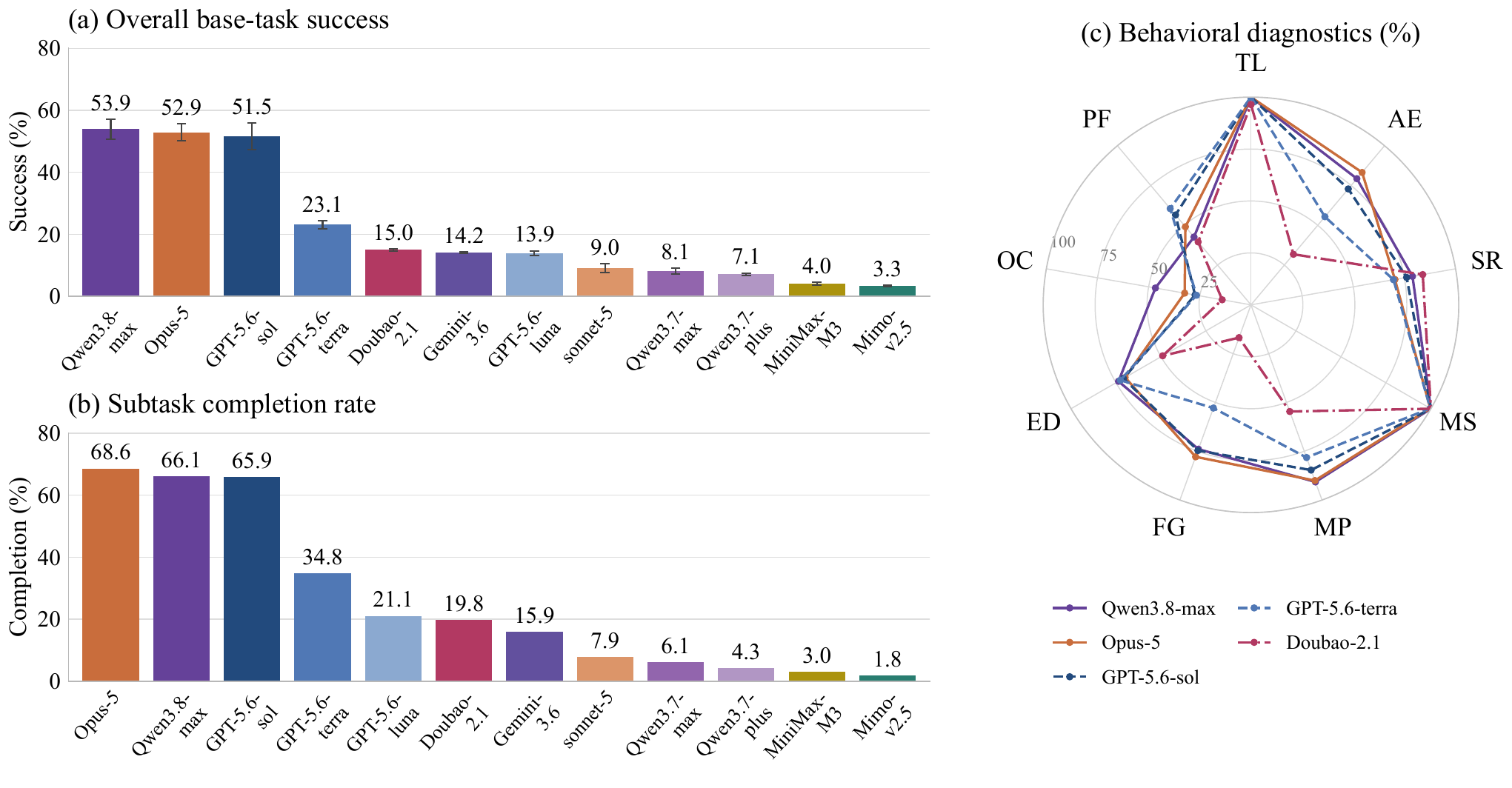}
    \caption{Execution success, subtask progress, and behavioral diagnostics.
    (a) Mean terminal success and sample standard deviation across three runs
    on 14 base-task families.  (b) Micro-averaged subtask completion from one
    annotated run per condition.  (c) Pooled majority-vote percentages for
    the five strongest conditions.  Panels (a) and (b) include all 12 models;
    Appendix Table~\ref{tab:capability-diagnostics} gives their full diagnostic
    results.}
    \label{fig:capability-radar}
\end{figure}

We evaluate 12 models using their own demonstration-derived summaries.
Terminal success is a task-family macro-average, reported as mean $\pm$
sample standard deviation across three runs.  Diagnostics and subtask progress
use one annotated run per model.  Appendix~\ref{app:base-evaluation-protocol}
specifies the evaluation conditions and aggregation; complete per-task and
arm-specific results appear in Appendix~\ref{app:complete-results}.

\subsection{Base-task performance}
\label{sec:core-results}

Table~\ref{tab:main-results} compares 12 primary model conditions on the 11
single-arm and three dual-arm tasks.  Qwen3.8-max has the highest three-run
mean at $53.93\!\pm\!3.17\%$, followed by Opus-5 at
$52.86\!\pm\!2.79\%$ and GPT-5.6-sol at $51.55\!\pm\!4.31\%$.  The top three
means lie within 2.38 percentage points, and their run-level ranges overlap.
With only three runs, we do not claim a reliable ordering among them.
Performance differs across the arm subsets.  Qwen3.8-max has the highest
single-arm mean ($65.61\%$), and GPT-5.6-sol has the highest dual-arm mean
($22.22\%$).

In the annotated run, the same three conditions lead micro-averaged subtask
completion and stay within 2.7 points of one another.  Opus-5 reaches
$68.57\%$, Qwen3.8-max $66.07\%$, and GPT-5.6-sol $65.89\%$.
Opus-5 makes the most intermediate progress, although its terminal success
is not distinguishable from the other two.  Progress uses checkpoint-level
weighting and is not an upper bound on task-family success.

Performance falls sharply below the top three.  GPT-5.6-terra reaches
$23.10\!\pm\!1.25\%$, and every remaining condition is at or below $15.00\%$.
Coordinated dual-arm execution remains difficult for all models.  Only the top
three conditions exceed a $10\%$ mean, and the best reaches just $22.22\%$.
Appendix
Tables~\ref{tab:appendix-core-a}--\ref{tab:appendix-core-c} provide the complete
task-by-condition matrix.

\begin{table}[H]
\caption{VA-Bench base-task success (\%) for 12 primary model conditions.
Entries are mean $\pm$ sample standard deviation across three runs; each run
macro-averages 11 single-arm tasks, three dual-arm tasks, or all 14 tasks.}
\label{tab:main-results}
\begin{center}
\begingroup
\scriptsize
\setlength{\tabcolsep}{2.0pt}
\renewcommand{\arraystretch}{0.95}
\begin{tabular}{@{}l*{6}{c}@{}}
\hline
Subset & \shortstack{Qwen3.8\\max} & \shortstack{Opus\\5} & \shortstack{GPT-5.6\\sol} &
\shortstack{GPT-5.6\\terra} & \shortstack{Doubao\\2.1} &
\shortstack{Gemini\\3.6} \\
\hline
Single-arm & \best{65.61$\pm$3.81} & 64.09$\pm$3.18 & 59.55$\pm$5.06 & 29.24$\pm$1.39 & 19.09$\pm$0.45 & 18.03$\pm$0.26 \\
Dual-arm   & 11.11$\pm$2.55 & 11.67$\pm$1.67 & \best{22.22$\pm$1.92} & 0.56$\pm$0.96 & 0.00$\pm$0.00 & 0.00$\pm$0.00 \\
All 14     & \best{53.93$\pm$3.17} & 52.86$\pm$2.79 & 51.55$\pm$4.31 & 23.10$\pm$1.25 & 15.00$\pm$0.36 & 14.17$\pm$0.21 \\
\hline
Subset & \shortstack{GPT-5.6\\luna} & \shortstack{sonnet\\5} &
\shortstack{Qwen3.7\\max} & \shortstack{Qwen3.7\\plus} &
\shortstack{MiniMax\\M3} & \shortstack{Mimo\\v2.5} \\
\hline
Single-arm & 17.42$\pm$0.69 & 11.52$\pm$1.89 & 10.30$\pm$1.14 & 9.09$\pm$0.45 & 5.15$\pm$0.69 & 4.24$\pm$0.26 \\
Dual-arm   & 1.11$\pm$0.96 & 0.00$\pm$0.00 & 0.00$\pm$0.00 & 0.00$\pm$0.00 & 0.00$\pm$0.00 & 0.00$\pm$0.00 \\
All 14     & 13.93$\pm$0.71 & 9.05$\pm$1.49 & 8.10$\pm$0.90 & 7.14$\pm$0.36 & 4.05$\pm$0.55 & 3.33$\pm$0.21 \\
\hline
\end{tabular}
\endgroup
\end{center}
\end{table}

\subsection{Interaction and failure analysis}
\label{sec:interaction-analysis}

Figure~\ref{fig:capability-radar} shows an uneven diagnostic profile.
The top three conditions score $100.0\%$ on target localization (TL) and at
least $99.6\%$ on manipulation semantics (MS).  Their spatial-relation scores
(SR) are lower: $78.9\%$ for Qwen3.8-max, $70.4\%$ for Opus-5, and $76.1\%$
for GPT-5.6-sol.  Identifying the target and selecting meaningful motions
does not establish accurate robot--object geometry.

Opus-5 has the highest active-exploration (AE) score at $83.2\%$ and the
highest fine-grained pre-contact analysis (FG) score at $77.9\%$.  GPT-5.6-sol and
Qwen3.8-max follow on FG at $74.6\%$ and $73.9\%$.  GPT-5.6-terra reaches
$52.9\%$, and every remaining condition is below $40\%$.
FG records analysis before contact, not successful operation completion.
These trajectory-level scores complement strict terminal success; they do not
replace it.  Appendix Table~\ref{tab:capability-diagnostics} reports all 12
conditions.

Recovery performance also varies by criterion.  Error detection is comparable
across Qwen3.8-max ($73.6\%$), GPT-5.6-terra ($72.6\%$), and Opus-5
($69.9\%$).  Online correction differs sharply: Qwen3.8-max reaches
$46.7\%$, Opus-5 $32.4\%$, and GPT-5.6-terra $26.6\%$.  Gemini-3.6
leads PF at $61.9\%$ despite near-zero OC.  High error-detection performance does not
imply that a model can correct the deviation online or replan after an explicit
failure.  The diagnostics record which behaviors occur and where they stop
within a trajectory, information that terminal success alone does not provide.

The three annotators agree unanimously on $95.7\%$ of applicable labels
(Fleiss $\kappa=0.942$).  Appendix~\ref{sec:app-annotation-reliability}
reports per-criterion agreement, sensitivity to the voting rule, and the
relationship between diagnostic labels and terminal outcomes.

\section{Generalization and Ablation}
\label{sec:generalization}

We first vary the camera observation protocol, then test four forms of
robustness beyond the base benchmark: controlled instance transfer,
long-horizon skill composition, transfer of a demonstration summary across
models, and summary-guided evaluation of locally deployed small models.  Seven
single-arm families vary geometry, container identity, or layout while reusing
the base-task summary.  A separate 20-episode track requires five-object
placement from four independently demonstrated atomic skills, without a
demonstration of the full trajectory.
These controlled analyses use their original matched single-run trajectories;
the three-run aggregation in Section~\ref{sec:main-results} applies to the
primary base-task terminal-success comparison.

\subsection{Active Evidence Acquisition vs. Passive Multi-View Input}
\label{sec:camera-protocol-ablation}

We compare the canonical active-view protocol with a passive multi-view
alternative for four conditions that span the observed performance range:
Qwen3.8-max, GPT-5.6-sol, Opus-5, and Doubao-2.1.  For each condition, the
passive run uses the setting of its first base-evaluation run.  We retain the
model checkpoint, its own demonstration summary, the same 14 task families
with the same 20 verified seeds each, and the same robot-action interface,
executor, action budget, terminal checker, and scoring.  The context limit and
compaction policy are also fixed.  Only the camera interface changes, so each
model is compared against itself.

The active condition gives the model one current RGB image and
semantic or local camera actions, which consume the shared episode budget.
The passive condition disables camera actions and supplies five predefined
views at every observation point; the model cannot select, reorder, or extend
this bundle.  The active camera interface can reach the same five canonical
viewpoints and additionally supports bounded incremental refinement.  The fixed
five-view bundle is therefore a strict subset of the viewpoints reachable in
active mode.

Each passive request contains only the current five-image bundle.  Prior robot
actions, planner outcomes, and textual feedback remain in history as in the
active condition.  This comparison tests model-directed decisions about whether
and where to acquire evidence, including local viewpoint refinement, against
predefined five-view coverage.  Appendix~\ref{sec:app-five-view-protocol} gives
the view definitions and payload order.  Table~\ref{tab:multiview-comparison}
reports pooled results.

\begin{table}[!htbp]
\centering
\caption{Controlled single-run active-versus-passive evidence-acquisition
comparison for four conditions.  Each condition reuses its own base-evaluation
summary, the same 20 seeds per family, and the same executor, budget, and
checker; only camera control differs.  Entries are success count / total
episodes followed by success rate, and $\Delta$ is passive minus active.}
\label{tab:multiview-comparison}
\small
\setlength{\tabcolsep}{4.0pt}
\begin{tabular}{llccc}
\hline
Model & Protocol & Single-arm (11$\times$20) & Dual-arm (3$\times$20) & All 14 (280) \\
\hline
Qwen3.8-max & Active  & 154/220 (70.00\%) & 7/60 (11.67\%) & 161/280 (57.50\%) \\
            & Passive & 77/220 (35.00\%) & 1/60 (1.67\%) & 78/280 (27.86\%) \\
            & $\Delta$ & $-35.00$ pp & $-10.00$ pp & $-29.64$ pp \\
\hline
GPT-5.6-sol & Active  & 143/220 (65.00\%) & 14/60 (23.33\%) & 157/280 (56.07\%) \\
            & Passive & 92/220 (41.82\%) & 3/60 (5.00\%) & 95/280 (33.93\%) \\
            & $\Delta$ & $-23.18$ pp & $-18.33$ pp & $-22.14$ pp \\
\hline
Opus-5     & Active  & 144/220 (65.45\%) & 8/60 (13.33\%) & 152/280 (54.29\%) \\
            & Passive & 82/220 (37.27\%) & 3/60 (5.00\%) & 85/280 (30.36\%) \\
            & $\Delta$ & $-28.18$ pp & $-8.33$ pp & $-23.93$ pp \\
\hline
Doubao-2.1 & Active  & 42/220 (19.09\%) & 0/60 (0.00\%) & 42/280 (15.00\%) \\
            & Passive & 15/220 (6.82\%) & 0/60 (0.00\%) & 15/280 (5.36\%) \\
            & $\Delta$ & $-12.27$ pp & $\pm0.00$ pp & $-9.64$ pp \\
\hline
\end{tabular}
\end{table}

Removing camera control reduces success by $29.64$, $22.14$, $23.93$, and
$9.64$ points, respectively.  No condition matches its active score with broader
passive coverage.  The effect spans the tested performance range, including the
two strongest conditions, a mid-range condition, and Doubao-2.1.  Even
its active score of $15.00\%$ falls by nearly two thirds in relative
terms.  Aggregating at the
task-family level, passive observation is worse on $10$ of $12$ non-tied families
for Qwen3.8-max, on all $12$ for GPT-5.6-sol and Opus-5, and on $8$ of $9$ for
Doubao-2.1 (two-sided sign tests, $p=0.039$, $0.0005$, $0.0005$, and $0.039$).
The degradation also concentrates on tasks that require task-conditioned
evidence.  Placement families show sharp declines under passive observation.
For Qwen3.8-max, \nolinkurl{place_cube_in_bowl} falls from $85.00\%$ to $5.00\%$
and \nolinkurl{place_single_cube} from $60.00\%$ to $10.00\%$.
Grasping families degrade far less; these depend on locating a single object
that the fixed views already cover.  Broader scene coverage does not replace
directing observation at the region relevant to the current subgoal.
The mechanism behind the gap remains open.  It may involve view relevance,
cross-view integration, or other factors
(Appendix~\ref{app:scope-limitations}).

\begin{table}[!htbp]
\centering
\caption{Controlled transfer and composition.  Base and held-out values average
seven matched families (20 seeds each); $\Delta$ is held-out minus base.  Long
reports successful object placements out of 100 possible and strict episode
success (\%) over 20 episodes.}
\label{tab:generalization-summary}
\small
\setlength{\tabcolsep}{4.0pt}
\begin{tabular}{lrrrcc}
\hline
& \multicolumn{3}{c}{Matched transfer (\%)} &
\multicolumn{2}{c}{Long horizon} \\
\cline{2-4}\cline{5-6}
Condition & Base & Held-out & $\Delta$ & Objects & Success \\
\hline
Qwen3.8-max   & 77.86 & \best{67.86} & $-10.00$ & \best{61/100} & \best{0.00} \\
GPT-5.6-sol   & \best{80.00} & 47.86 & $-32.14$ & 53/100 & \best{0.00} \\
GPT-5.6-terra & 35.00 & 27.86 &  $-7.14$ & 5/100 & \best{0.00} \\
GPT-5.6-luna  & 22.86 & 18.57 &  \best{$-4.29$} & 2/100 & \best{0.00} \\
\hline
\end{tabular}
\end{table}

\paragraph{Controlled transfer.}
Qwen3.8-max transfers best, retaining $67.86\%$ success with a $10.00$-point
drop.  GPT-5.6-sol starts from a comparable matched-base rate but loses
$32.14$ points.  Strong base execution thus does not ensure robust spatial
re-grounding.  The smaller drops of weaker conditions occur at much lower
performance floors.  Appendix Table~\ref{tab:single-arm-generalization}
reports all seven families.

\paragraph{Long-horizon composition.}
No model completes a strict long-horizon episode.  Qwen3.8-max achieves the
greatest partial progress with 61/100 successful object placements, followed by
GPT-5.6-sol with 53/100.  This partial progress is insufficient for complete
inventory-conditioned composition: strict task success remains zero.

\paragraph{Trajectory diagnostics.}
Under controlled transfer, spatial perception and understanding remains at or
above $99.3\%$ for all four conditions.  Robot manipulation ranges from
$54.3\%$ to $86.7\%$.  High spatial-perception module scores coexist with
lower manipulation scores and incomplete task execution.  Appendix
Table~\ref{tab:generalization-capabilities} reports the three descriptive
module means.

\subsection{Demonstration-summary ablation}
\label{sec:summary-source}

A matched Qwen3.7-plus comparison holds the evaluated model, task set, and
seeds fixed.  Replacing its model-specific summary with the Qwen3.8-max summary
reduces the 14-task macro-average from $7.14\%$ to $5.00\%$ and leaves
dual-arm success at zero.  A stronger model's textual summary is not
automatically a better interface for another model.  Appendix
Table~\ref{tab:summary-source-ablation} gives the subset breakdown.  This
comparison covers one reader, so we next test several readers.

\paragraph{Cross-model summary sources.}
VA-Bench primarily measures how a model understands a task, learns it from a
demonstration, and integrates spatial understanding with manipulation.  These
results combine the quality of the written interface a model produces with its
ability to act on that interface.  We hold the interface fixed and vary the
reader to examine these factors.  Shared base summaries come from the three
strongest conditions: Qwen3.8-max, GPT-5.6-sol, and Opus-5.  We evaluate five
models under each summary source on three tasks (\texttt{grasp\_single\_bottle},
\texttt{grasp\_single\_pen}, \texttt{place\_cube\_in\_bowl}).  For each model,
all settings except the summary match its first base-evaluation run.  This
includes the same 20 verified seeds per task, executor, action budget, and
terminal checker.  Each cell contains 60 episodes.  The \emph{Own} column of
Table~\ref{tab:summary-cross-eval} reports that base run.  For the three summary
authors, their own-summary cells reproduce it identically, making the shared
columns directly comparable to this baseline.

\begin{table}[!htbp]
\caption{Shared demonstration-summary sources.  Rates are pooled success (\%)
over three tasks with 20 seeds each; \emph{Own} is the model's own
model-specific summary in its first base-evaluation run, and $\Delta$ is
shared minus own.  In the \emph{avg} row, rates average all five models;
$\Delta$ excludes the summary author, whose own-summary cell is its base run
by construction.}
\label{tab:summary-cross-eval}
\begin{center}
\small
\setlength{\tabcolsep}{4.0pt}
\begin{tabular}{lrrrrrrr}
\hline
& & \multicolumn{2}{c}{On Qwen3.8-max} & \multicolumn{2}{c}{On GPT-5.6-sol} &
\multicolumn{2}{c}{On Opus-5} \\
\cline{3-4}\cline{5-6}\cline{7-8}
Evaluated model & Own & Rate & $\Delta$ & Rate & $\Delta$ & Rate & $\Delta$ \\
\hline
Qwen3.8-max   & \best{86.67} & \best{86.67} &   $0.00$ & 85.00 &  $-1.67$ & \best{88.33} &  $+1.67$ \\
Opus-5        & 85.00 & 56.67 & $-28.33$ & 78.33 &  $-6.67$ & 85.00 &   $0.00$ \\
GPT-5.6-sol   & 73.33 & 21.67 & $-51.67$ & 73.33 &   $0.00$ & 76.67 &  $+3.33$ \\
GPT-5.6-terra & 41.67 & 10.00 & $-31.67$ & 41.67 &   $0.00$ & 55.00 & $+13.33$ \\
Gemini-3.6    & 23.33 & 10.00 & $-13.33$ & 35.00 & $+11.67$ & 35.00 & $+11.67$ \\
\hline
avg                  & 62.00 & 37.00 & $-31.25$ & 62.67 & $+0.83$ & \best{68.00} & \best{$+7.50$} \\
\hline
\end{tabular}
\end{center}
\end{table}

The model with the highest task completion does not produce the most
transferable summary.  Qwen3.8-max leads task completion, but its summary lowers
success for all four other models by $31.25$ points on average.  The largest
drop is $51.67$ points for GPT-5.6-sol.  The Opus-5 summary improves all four
other models by $7.50$ points on average.  GPT-5.6-sol's summary is roughly
neutral at $+0.83$.

The content of the summaries suggests a possible explanation.  Qwen3.8-max
states the fewest constraints and leaves object-specific detail implicit.
GPT-5.6-sol and Opus-5 explicitly constrain both fine-grained details and the
overall procedure; Opus-5 is the most explicit of the three.  A permissive
summary may be sufficient for its author but underspecified for the other
models.  More explicit constraints may allow a summary to transfer with little
loss.  They may also help the weakest readers most by supplying details that
those models omit from their own summaries.

Under the Opus-5 summary, Gemini-3.6 rises from $23.33\%$ to $35.00\%$ and
GPT-5.6-terra from $41.67\%$ to $55.00\%$.  The gain is concentrated on the
placement task that neither completes reliably on its own
(\texttt{place\_cube\_in\_bowl}: $0\%\rightarrow20\%$ and
$15\%\rightarrow40\%$).  These observations suggest that summary transfer
depends in part on how explicitly the constraints are stated, not on the task
success of the authoring model.  The comparison covers three tasks in a single
run.  We therefore treat the ordering of the sources as a suggestive trend
rather than a precise estimate of transfer effects.  Appendix
Table~\ref{tab:summary-cross-eval-tasks} gives the per-task breakdown.

\subsection{Summary-guided local small-model evaluation}

We further evaluate four locally deployed models using summaries sourced from
Qwen3.8-max.  Qwen3.8-27B and Gemma-4-31B-it obtain 27/280 ($9.64\%$) and
24/280 ($8.57\%$), while the two Qwen3.6 variants reach 3/280 ($1.07\%$) and
2/280 ($0.71\%$).  For Gemma-4-31B-it and the Qwen3.6 variants, the corresponding
unguided controls obtain zero success.  Summary-guided successes remain confined
to grasping and bell interaction: no condition completes a placement or
dual-arm task.  Appendix Tables~\ref{tab:local-small-model-summary}
and~\ref{tab:local-small-model-tasks} provide the aggregate and task-level
breakdowns.

\section{Conclusion}
\label{sec:conclusion}

VA-Bench evaluates whether a general-purpose MLLM can close the active
observe--reason--act loop through metric robot commands, without target poses,
oracle trajectories, or a learned action head.  Across 14 task families and
three runs, the best macro-average is $53.93\!\pm\!3.17\%$.  Models achieve
strong target-localization and manipulation-semantics scores, but
spatial-relation diagnostics are less consistent and online correction remains
weak.

Strong target-localization scores do not ensure precise, composable, and
recoverable physical action.  Strict environment success measures whether a
model completes the task.  The trajectory diagnostics show where models stop
using spatial evidence effectively during execution.

In separate matched protocol runs, removing camera control reduces terminal
success by $9.64$ to $29.64$ percentage points across the tested conditions.
Qwen3.8-max, for example, falls from $57.50\%$ to $27.86\%$.  GPT-5.6-sol,
Opus-5, and Doubao-2.1 show the same direction.  Broader passive coverage does
not substitute for task-conditioned evidence acquisition.

\clearpage
\bibliography{va_bench_references_final}
\bibliographystyle{iclr2027_conference}

\clearpage
\appendix
\section{Complete Evaluation Results}
\label{app:complete-results}

\setcounter{topnumber}{5}
\setcounter{bottomnumber}{5}
\setcounter{totalnumber}{8}
\renewcommand{\topfraction}{0.95}
\renewcommand{\bottomfraction}{0.95}
\renewcommand{\textfraction}{0.05}

This appendix reports the disaggregated values underlying the compact tables
in the main paper.  All tables remain in the normal reading orientation.  The
per-task base-benchmark matrix is split by model group, and the nine
behavioral diagnostics are split by module.  Model columns and rows follow descending
14-task macro-average; bold blue cells mark the best score within each directly
comparable task row or metric column, including ties.

\subsection{Base-evaluation settings and aggregation}
\label{app:base-evaluation-protocol}

\paragraph{Conditions and validity.}
The primary comparison evaluates 12 models, each using a summary it produced
from the task's RGB-only expert video.  Two shared-summary controls reuse
another model's summary and are reported separately in
Section~\ref{sec:generalization}: one tests summary-source transfer and the
other supports the local-model comparison.  Their outcomes are not pooled
with the primary conditions.

Following Section~\ref{sec:benchmark-construction}, each independent run uses
the same 20 retained seeds for each of the 11 single-arm and three dual-arm
families.  Three runs give 60 outcomes per task and 840 run--seed outcomes per
primary condition.  Offline verification signals remain evaluator-only, and
terminal success is determined solely by the environment checker.  All models
share the observation, action, budget, and executor contract in
Section~\ref{sec:benchmark}; harness details appear in
Appendix~\ref{app:agent-harness-details}.

\paragraph{Terminal success.}
Let $y_{m,r,\tau,i}\in\{0,1\}$ be the checker outcome for model $m$, run $r$,
task $\tau$, and seed $i$.  We average over seeds within a task and then
macro-average over task families:
\begin{equation}
S_{m,r,\tau}=\frac{1}{20}\sum_{i=1}^{20}y_{m,r,\tau,i},
\qquad
S_{m,r}^{\mathrm{all}}=\frac{1}{14}\sum_{\tau=1}^{14}S_{m,r,\tau}.
\label{eq:base-success}
\end{equation}
We report the mean and sample standard deviation of the three run-level
macro-averages.  Single- and dual-arm scores apply the same procedure over
11 and three families, respectively.  Every family has equal weight, and the
error bars measure run-to-run variation, not binomial uncertainty over seeds.

\paragraph{Annotated run and partial progress.}
Diagnostics and subtask progress use one complete base run per primary
condition, totaling 3,360 episodes.  Eleven conditions use their first run;
sonnet-5 uses its second.  Each diagnostic pools applicable episode-level
majority-vote labels across the single- and dual-arm subsets.  Figure~\ref{fig:capability-radar}
plots these percentages directly, without averaging individual annotator votes
or applying min--max normalization.

Subtask completion is a micro-average over scored checkpoint instances, with
420 single-arm and 140 dual-arm checkpoints per model.  Because its weighting
unit is a checkpoint rather than a task family, it measures intermediate
progress but is not a relaxed upper bound on terminal success.
Appendix~\ref{sec:app-diagnostic-scoring} defines observable checkpoint evidence
and diagnostic applicability.  Appendix~\ref{sec:app-annotation-reliability}
reports agreement and sensitivity analyses.

\subsection{Base-task success rates}

Tables~\ref{tab:appendix-core-a}--\ref{tab:appendix-core-c} report the
three-run mean and sample standard deviation for each of the 14 base tasks and
12 primary model conditions.  Each run evaluates the same 20 retained
seeds meeting the acceptance criteria in Appendix~\ref{sec:app-seed-validation};
the final row aggregates each run over all 14 tasks before computing summary
statistics.  Steps is the maximum interaction horizon.  Doubao-2.1 and
Gemini-3.6 abbreviate \texttt{doubao-seed-2.1-turbo} and
\texttt{gemini-3.6-flash}, respectively.

\begin{table}[htbp]
\caption{Base-task success rates (\%, mean $\pm$ sample SD), model group 1 of 3.}
\label{tab:appendix-core-a}
\centering
\begingroup
\scriptsize
\setlength{\tabcolsep}{1.8pt}
\renewcommand{\arraystretch}{0.92}
\begin{tabular}{@{}p{2.05in}r*{4}{r}@{}}
\hline
Task & Steps & \shortstack{Qwen3.8\\max} & \shortstack{Opus\\5} &
\shortstack{GPT-5.6\\sol} & \shortstack{GPT-5.6\\terra} \\
\hline
\multicolumn{6}{@{}l}{\textit{Single-arm tasks}} \\
\nolinkurl{beat_block_hammer_right} & 100 & 3.33$\pm$2.89 & 8.33$\pm$2.89 & \best{15.00$\pm$5.00} & 6.67$\pm$2.89 \\
\nolinkurl{click_bell_right} & 50 & \best{73.33$\pm$5.77} & 68.33$\pm$14.43 & 65.00$\pm$15.00 & 1.67$\pm$2.89 \\
\nolinkurl{grasp_pen_leaning_cube} & 50 & \best{61.67$\pm$2.89} & 35.00$\pm$5.00 & 45.00$\pm$18.03 & 25.00$\pm$5.00 \\
\nolinkurl{grasp_single_bottle} & 50 & \best{91.67$\pm$2.89} & 75.00$\pm$0.00 & 76.67$\pm$11.55 & 51.67$\pm$5.77 \\
\nolinkurl{grasp_single_bottle_upright} & 50 & 65.00$\pm$5.00 & 78.33$\pm$16.07 & 70.00$\pm$10.00 & 68.33$\pm$2.89 \\
\nolinkurl{grasp_single_cube} & 50 & \best{83.33$\pm$10.41} & \best{83.33$\pm$11.55} & 76.67$\pm$17.56 & 31.67$\pm$5.77 \\
\nolinkurl{grasp_single_pen} & 50 & \best{91.67$\pm$10.41} & 63.33$\pm$15.28 & 28.33$\pm$16.07 & 48.33$\pm$7.64 \\
\nolinkurl{place_cube_in_bowl} & 50 & 81.67$\pm$2.89 & \best{88.33$\pm$10.41} & 85.00$\pm$10.00 & 18.33$\pm$2.89 \\
\nolinkurl{place_cube_on_cube} & 200 & 86.67$\pm$7.64 & 78.33$\pm$2.89 & \best{88.33$\pm$12.58} & 46.67$\pm$2.89 \\
\nolinkurl{place_single_bottle_upright} & 100 & \best{40.00$\pm$5.00} & 38.33$\pm$2.89 & 13.33$\pm$2.89 & 6.67$\pm$5.77 \\
\nolinkurl{place_single_cube} & 100 & 43.33$\pm$14.43 & 88.33$\pm$7.64 & \best{91.67$\pm$7.64} & 16.67$\pm$2.89 \\
\multicolumn{6}{@{}l}{\textit{Dual-arm tasks}} \\
\nolinkurl{place_shoe} & 100 & 33.33$\pm$7.64 & 35.00$\pm$5.00 & \best{66.67$\pm$5.77} & 0.00$\pm$0.00 \\
\nolinkurl{handover_horizontal_block} & 100 & \best{0.00$\pm$0.00} & \best{0.00$\pm$0.00} & \best{0.00$\pm$0.00} & \best{0.00$\pm$0.00} \\
\nolinkurl{lift_pot} & 50 & 0.00$\pm$0.00 & 0.00$\pm$0.00 & 0.00$\pm$0.00 & \best{1.67$\pm$2.89} \\
\hline
\textit{Macro-average} & & \best{53.93$\pm$3.17} & 52.86$\pm$2.79 & 51.55$\pm$4.31 & 23.10$\pm$1.25 \\
\hline
\end{tabular}
\endgroup
\end{table}

\begin{table}[htbp]
\caption{Base-task success rates (\%, mean $\pm$ sample SD), model group 2 of 3.}
\label{tab:appendix-core-b}
\centering
\begingroup
\scriptsize
\setlength{\tabcolsep}{1.8pt}
\renewcommand{\arraystretch}{0.92}
\begin{tabular}{@{}p{2.05in}r*{4}{r}@{}}
\hline
Task & Steps & \shortstack{Doubao\\2.1} & \shortstack{Gemini\\3.6} &
\shortstack{GPT-5.6\\luna} & \shortstack{sonnet\\5} \\
\hline
\multicolumn{6}{@{}l}{\textit{Single-arm tasks}} \\
\nolinkurl{beat_block_hammer_right} & 100 & 0.00$\pm$0.00 & 0.00$\pm$0.00 & 0.00$\pm$0.00 & 0.00$\pm$0.00 \\
\nolinkurl{click_bell_right} & 50 & 6.67$\pm$5.77 & 13.33$\pm$2.89 & 0.00$\pm$0.00 & 11.67$\pm$2.89 \\
\nolinkurl{grasp_pen_leaning_cube} & 50 & 6.67$\pm$2.89 & 10.00$\pm$5.00 & 18.33$\pm$2.89 & 5.00$\pm$8.66 \\
\nolinkurl{grasp_single_bottle} & 50 & 86.67$\pm$2.89 & 38.33$\pm$2.89 & 3.33$\pm$5.77 & 33.33$\pm$5.77 \\
\nolinkurl{grasp_single_bottle_upright} & 50 & 48.33$\pm$2.89 & 78.33$\pm$2.89 & \best{86.67$\pm$5.77} & 40.00$\pm$10.00 \\
\nolinkurl{grasp_single_cube} & 50 & 10.00$\pm$5.00 & 8.33$\pm$2.89 & 15.00$\pm$5.00 & 16.67$\pm$2.89 \\
\nolinkurl{grasp_single_pen} & 50 & 26.67$\pm$10.41 & 23.33$\pm$7.64 & 8.33$\pm$2.89 & 1.67$\pm$2.89 \\
\nolinkurl{place_cube_in_bowl} & 50 & 6.67$\pm$7.64 & 5.00$\pm$5.00 & 6.67$\pm$7.64 & 11.67$\pm$2.89 \\
\nolinkurl{place_cube_on_cube} & 200 & 8.33$\pm$2.89 & 10.00$\pm$5.00 & 13.33$\pm$2.89 & 0.00$\pm$0.00 \\
\nolinkurl{place_single_bottle_upright} & 100 & 6.67$\pm$2.89 & 11.67$\pm$2.89 & 26.67$\pm$7.64 & 0.00$\pm$0.00 \\
\nolinkurl{place_single_cube} & 100 & 3.33$\pm$2.89 & 0.00$\pm$0.00 & 13.33$\pm$2.89 & 6.67$\pm$2.89 \\
\multicolumn{6}{@{}l}{\textit{Dual-arm tasks}} \\
\nolinkurl{place_shoe} & 100 & 0.00$\pm$0.00 & 0.00$\pm$0.00 & 3.33$\pm$2.89 & 0.00$\pm$0.00 \\
\nolinkurl{handover_horizontal_block} & 100 & \best{0.00$\pm$0.00} & \best{0.00$\pm$0.00} & \best{0.00$\pm$0.00} & \best{0.00$\pm$0.00} \\
\nolinkurl{lift_pot} & 50 & 0.00$\pm$0.00 & 0.00$\pm$0.00 & 0.00$\pm$0.00 & 0.00$\pm$0.00 \\
\hline
\textit{Macro-average} & & 15.00$\pm$0.36 & 14.17$\pm$0.21 & 13.93$\pm$0.71 & 9.05$\pm$1.49 \\
\hline
\end{tabular}
\endgroup
\end{table}

\begin{table}[H]
\caption{Base-task success rates (\%, mean $\pm$ sample SD), model group 3 of 3.}
\label{tab:appendix-core-c}
\centering
\begingroup
\scriptsize
\setlength{\tabcolsep}{1.8pt}
\renewcommand{\arraystretch}{0.92}
\begin{tabular}{@{}p{2.05in}r*{4}{r}@{}}
\hline
Task & Steps & \shortstack{Qwen3.7\\max} & \shortstack{Qwen3.7\\plus} &
\shortstack{MiniMax\\M3} & \shortstack{Mimo\\v2.5} \\
\hline
\multicolumn{6}{@{}l}{\textit{Single-arm tasks}} \\
\nolinkurl{beat_block_hammer_right} & 100 & 0.00$\pm$0.00 & 0.00$\pm$0.00 & 0.00$\pm$0.00 & 0.00$\pm$0.00 \\
\nolinkurl{click_bell_right} & 50 & 0.00$\pm$0.00 & 0.00$\pm$0.00 & 0.00$\pm$0.00 & 0.00$\pm$0.00 \\
\nolinkurl{grasp_pen_leaning_cube} & 50 & 1.67$\pm$2.89 & 3.33$\pm$2.89 & 0.00$\pm$0.00 & 0.00$\pm$0.00 \\
\nolinkurl{grasp_single_bottle} & 50 & 1.67$\pm$2.89 & 18.33$\pm$2.89 & 0.00$\pm$0.00 & 0.00$\pm$0.00 \\
\nolinkurl{grasp_single_bottle_upright} & 50 & 76.67$\pm$7.64 & 76.67$\pm$2.89 & 46.67$\pm$2.89 & 41.67$\pm$10.41 \\
\nolinkurl{grasp_single_cube} & 50 & 20.00$\pm$10.00 & 0.00$\pm$0.00 & 10.00$\pm$5.00 & 5.00$\pm$8.66 \\
\nolinkurl{grasp_single_pen} & 50 & 13.33$\pm$2.89 & 0.00$\pm$0.00 & 0.00$\pm$0.00 & 0.00$\pm$0.00 \\
\nolinkurl{place_cube_in_bowl} & 50 & 0.00$\pm$0.00 & 1.67$\pm$2.89 & 0.00$\pm$0.00 & 0.00$\pm$0.00 \\
\nolinkurl{place_cube_on_cube} & 200 & 0.00$\pm$0.00 & 0.00$\pm$0.00 & 0.00$\pm$0.00 & 0.00$\pm$0.00 \\
\nolinkurl{place_single_bottle_upright} & 100 & 0.00$\pm$0.00 & 0.00$\pm$0.00 & 0.00$\pm$0.00 & 0.00$\pm$0.00 \\
\nolinkurl{place_single_cube} & 100 & 0.00$\pm$0.00 & 0.00$\pm$0.00 & 0.00$\pm$0.00 & 0.00$\pm$0.00 \\
\multicolumn{6}{@{}l}{\textit{Dual-arm tasks}} \\
\nolinkurl{place_shoe} & 100 & 0.00$\pm$0.00 & 0.00$\pm$0.00 & 0.00$\pm$0.00 & 0.00$\pm$0.00 \\
\nolinkurl{handover_horizontal_block} & 100 & \best{0.00$\pm$0.00} & \best{0.00$\pm$0.00} & \best{0.00$\pm$0.00} & \best{0.00$\pm$0.00} \\
\nolinkurl{lift_pot} & 50 & 0.00$\pm$0.00 & 0.00$\pm$0.00 & 0.00$\pm$0.00 & 0.00$\pm$0.00 \\
\hline
\textit{Macro-average} & & 8.10$\pm$0.90 & 7.14$\pm$0.36 & 4.05$\pm$0.55 & 3.33$\pm$0.21 \\
\hline
\end{tabular}
\endgroup
\end{table}

\begin{table}[htbp]
\caption{Per-task successes out of 20 seeds for the controlled active evidence
acquisition versus passive five-view observation comparison, first
base-evaluation run.  Each condition reuses its own demonstration summary and
the same seeds in both protocols; robot execution and terminal success/failure
accounting are identical.  The active protocol permits camera actions, whereas
the passive protocol supplies five predefined views without camera control.
A/P denotes active/passive.}
\label{tab:multiview-per-task}
\begin{center}
\begingroup
\scriptsize
\setlength{\tabcolsep}{3.0pt}
\begin{tabular}{@{}p{1.72in}cccccccc@{}}
\hline
& \multicolumn{2}{c}{Qwen3.8-max} & \multicolumn{2}{c}{GPT-5.6-sol} &
\multicolumn{2}{c}{Opus-5} & \multicolumn{2}{c}{Doubao-2.1} \\
\cline{2-3}\cline{4-5}\cline{6-7}\cline{8-9}
Task & A & P & A & P & A & P & A & P \\
\hline
\multicolumn{9}{@{}l}{\textit{Single-arm tasks}} \\
\nolinkurl{click_bell_right}            & 16 & 19 & 16 & 14 & 12 & 10 & 0 & 3 \\
\nolinkurl{grasp_pen_leaning_cube}      & 13 & 11 & 5 & 2 & 8 & 6 & 1 & 0 \\
\nolinkurl{grasp_single_bottle}         & 19 & 12 & 18 & 11 & 15 & 13 & 17 & 7 \\
\nolinkurl{grasp_single_bottle_upright} & 14 & 8 & 16 & 13 & 12 & 7 & 10 & 2 \\
\nolinkurl{grasp_single_cube}           & 19 & 7 & 19 & 15 & 18 & 11 & 3 & 0 \\
\nolinkurl{grasp_single_pen}            & 16 & 11 & 7 & 6 & 16 & 13 & 7 & 3 \\
\nolinkurl{place_cube_in_bowl}          & 17 & 1 & 19 & 17 & 20 & 8 & 0 & 0 \\
\nolinkurl{place_cube_on_cube}          & 19 & 5 & 20 & 14 & 16 & 6 & 2 & 0 \\
\nolinkurl{place_single_bottle_upright} & 9 & 0 & 3 & 0 & 8 & 2 & 1 & 0 \\
\nolinkurl{place_single_cube}           & 12 & 2 & 17 & 0 & 18 & 6 & 1 & 0 \\
\nolinkurl{beat_block_hammer_right}     & 0 & 1 & 3 & 0 & 1 & 0 & 0 & 0 \\
\hline
\textit{Single-arm total (/220)} & 154 & 77 & 143 & 92 & 144 & 82 & 42 & 15 \\
\hline
\multicolumn{9}{@{}l}{\textit{Dual-arm tasks}} \\
\nolinkurl{place_shoe}                  & 7 & 1 & 14 & 3 & 8 & 3 & 0 & 0 \\
\nolinkurl{handover_horizontal_block}   & 0 & 0 & 0 & 0 & 0 & 0 & 0 & 0 \\
\nolinkurl{lift_pot}                    & 0 & 0 & 0 & 0 & 0 & 0 & 0 & 0 \\
\hline
\textit{Dual-arm total (/60)} & 7 & 1 & 14 & 3 & 8 & 3 & 0 & 0 \\
\textit{All-task total (/280)} & 161 & 78 & 157 & 95 & 152 & 85 & 42 & 15 \\
\hline
\end{tabular}
\endgroup
\end{center}
\end{table}

\subsection{Base-task behavioral diagnostics}

Tables~\ref{tab:appendix-core-spu}--\ref{tab:appendix-core-er} give the
single- and dual-arm components behind the pooled scores in
Table~\ref{tab:capability-diagnostics}.  Each pooled value combines valid
counts from the two subsets.  All four tables report the same twelve primary
conditions, each with one fully annotated run.  For eleven conditions the
annotated run is the first run; for sonnet-5 it is the second
($13.64\%$ single-arm success).  Every component-table entry is
a percentage.  TL: target
perception and localization; AE: informative active exploration; SR:
robot--object spatial relations; MS:
manipulation-semantics understanding; MP: goal-directed manipulation planning;
FG: fine-grained pre-contact analysis, scored over every reported episode; ED:
error detection; OC: online correction; PF: post-failure adjustment.
Annotators marked an assessable ambiguity in every base-task episode, so AE is
applicable to all 280 annotated episodes per condition in this track; the error-recovery
criteria retain their majority-voted applicability gate.
For ED, OC, and PF, an episode with no observable error is recorded as NA and
excluded from the denominator.  When an error occurs, all three criteria are
applicable and receive an explicit positive or negative judgment.

\begin{table}[htbp]
\caption{Pooled base-task trajectory-level behavioral diagnostics
(\%) for all twelve primary conditions.  Scores combine applicable single-
and dual-arm episodes; error-free episodes are excluded from ED, OC, and PF.}
\label{tab:capability-diagnostics}
\begin{center}
\begingroup
\scriptsize
\setlength{\tabcolsep}{2.0pt}
\renewcommand{\arraystretch}{0.95}
\begin{tabular}{@{}lrrrrrrrrr@{}}
\hline
Condition & TL & AE & SR & MS & MP & FG & ED & OC & PF \\
\hline
Qwen3.8-max & \best{100.0} & 79.3 & 78.9 & \best{100.0} & \best{90.7} & 73.9 & \best{73.6} & \best{46.7} & 42.6 \\
Opus-5 & \best{100.0} & \best{83.2} & 70.4 & 99.6 & 90.0 & \best{77.9} & 69.9 & 32.4 & 49.1 \\
GPT-5.6-sol & \best{100.0} & 72.9 & 76.1 & \best{100.0} & 84.6 & 74.6 & 70.8 & 27.3 & 56.5 \\
GPT-5.6-terra & \best{100.0} & 55.4 & 69.6 & 99.6 & 78.2 & 52.9 & 72.6 & 26.6 & 60.5 \\
Doubao-2.1 & 96.4 & 31.8 & 83.9 & \best{100.0} & 54.6 & 16.8 & 49.0 & 14.1 & 39.5 \\
Gemini-3.6 & \best{100.0} & 31.4 & 85.4 & \best{100.0} & 73.2 & 8.9 & 63.8 & 1.9 & \best{61.9} \\
GPT-5.6-luna & \best{100.0} & 40.0 & 71.8 & 99.3 & 61.1 & 38.9 & 49.2 & 13.7 & 37.4 \\
sonnet-5 & 99.3 & 21.8 & 80.7 & \best{100.0} & 31.8 & 32.1 & 10.6 & 1.2 & 9.8 \\
Qwen3.7-max & 96.4 & 13.6 & 81.4 & 97.5 & 35.7 & 2.5 & 32.6 & 0.4 & 32.2 \\
Qwen3.7-plus & 97.9 & 10.7 & 82.9 & 94.6 & 36.8 & 2.9 & 17.7 & 1.1 & 17.0 \\
MiniMax-M3 & 95.7 & 6.1 & \best{93.6} & 95.4 & 21.4 & 1.8 & 22.6 & 1.4 & 21.1 \\
Mimo-v2.5 & 83.2 & 5.4 & 71.8 & 98.9 & 19.6 & 2.1 & 25.8 & 0.0 & 26.2 \\
\hline
\end{tabular}
\endgroup
\end{center}
\end{table}

\begin{table}[ht]
\caption{Base-task spatial perception and understanding diagnostics
(\%) for all twelve primary conditions.}
\label{tab:appendix-core-spu}
\begin{center}
\begingroup
\scriptsize
\setlength{\tabcolsep}{2.8pt}
\begin{tabular}{lccc|ccc}
\hline
& \multicolumn{3}{c|}{Single-arm} & \multicolumn{3}{c}{Dual-arm} \\
Condition & TL & AE & SR & TL & AE & SR \\
\hline
Qwen3.8-max & \best{100.0} & 85.0 & 89.1 & \best{100.0} & \best{58.3} & 41.7 \\
Opus-5 & \best{100.0} & \best{90.5} & 71.4 & \best{100.0} & 56.7 & 66.7 \\
GPT-5.6-sol & \best{100.0} & 81.8 & 81.8 & \best{100.0} & 40.0 & 55.0 \\
GPT-5.6-terra & \best{100.0} & 64.1 & 70.0 & \best{100.0} & 23.3 & 68.3 \\
Doubao-2.1 & 95.5 & 34.1 & 85.5 & \best{100.0} & 23.3 & 78.3 \\
Gemini-3.6 & \best{100.0} & 36.8 & 89.1 & \best{100.0} & 11.7 & 71.7 \\
GPT-5.6-luna & \best{100.0} & 47.7 & 69.5 & \best{100.0} & 11.7 & 80.0 \\
sonnet-5 & 99.5 & 25.5 & 76.8 & 98.3 & 8.3 & \best{95.0} \\
Qwen3.7-max & 99.1 & 16.4 & 83.6 & 86.7 & 3.3 & 73.3 \\
Qwen3.7-plus & 99.1 & 13.2 & 84.1 & 93.3 & 1.7 & 78.3 \\
MiniMax-M3 & 96.4 & 7.3 & \best{94.1} & 93.3 & 1.7 & 91.7 \\
Mimo-v2.5 & 82.7 & 5.5 & 69.5 & 85.0 & 5.0 & 80.0 \\
\hline
\end{tabular}
\endgroup
\end{center}
\end{table}

\begin{table}[htbp]
\caption{Base-task robot manipulation diagnostics (\%) for all twelve
primary conditions.}
\label{tab:appendix-core-rm}
\begin{center}
\begingroup
\scriptsize
\setlength{\tabcolsep}{2.8pt}
\begin{tabular}{lccc|ccc}
\hline
& \multicolumn{3}{c|}{Single-arm} & \multicolumn{3}{c}{Dual-arm} \\
Condition & MS & MP & FG & MS & MP & FG \\
\hline
Qwen3.8-max & \best{100.0} & 94.1 & 75.5 & \best{100.0} & \best{78.3} & \best{68.3} \\
Opus-5 & 99.5 & \best{94.5} & \best{86.4} & \best{100.0} & 73.3 & 46.7 \\
GPT-5.6-sol & \best{100.0} & 90.0 & 76.4 & \best{100.0} & 65.0 & \best{68.3} \\
GPT-5.6-terra & 99.5 & 86.4 & 58.2 & \best{100.0} & 48.3 & 33.3 \\
Doubao-2.1 & \best{100.0} & 61.4 & 15.0 & \best{100.0} & 30.0 & 23.3 \\
Gemini-3.6 & \best{100.0} & 78.2 & 8.6 & \best{100.0} & 55.0 & 10.0 \\
GPT-5.6-luna & 99.1 & 63.2 & 45.0 & \best{100.0} & 53.3 & 16.7 \\
sonnet-5 & \best{100.0} & 33.6 & 40.0 & \best{100.0} & 25.0 & 3.3 \\
Qwen3.7-max & 96.8 & 43.2 & 3.2 & \best{100.0} & 8.3 & 0.0 \\
Qwen3.7-plus & 93.2 & 43.6 & 3.6 & \best{100.0} & 11.7 & 0.0 \\
MiniMax-M3 & 94.1 & 21.8 & 2.3 & \best{100.0} & 20.0 & 0.0 \\
Mimo-v2.5 & 98.6 & 25.0 & 2.7 & \best{100.0} & 0.0 & 0.0 \\
\hline
\end{tabular}
\endgroup
\end{center}
\end{table}

\begin{table}[H]
\caption{Base-task error recovery diagnostics (\%) for all twelve primary
conditions.  Error-free episodes are NA and excluded before computing each
percentage.}
\label{tab:appendix-core-er}
\begin{center}
\begingroup
\scriptsize
\setlength{\tabcolsep}{2.8pt}
\begin{tabular}{lccc|ccc}
\hline
& \multicolumn{3}{c|}{Single-arm} & \multicolumn{3}{c}{Dual-arm} \\
Condition & ED & OC & PF & ED & OC & PF \\
\hline
Qwen3.8-max & \best{81.7} & \best{54.2} & 43.0 & 52.7 & \best{27.3} & 41.8 \\
Opus-5 & 79.4 & 36.9 & 55.6 & 42.9 & 19.6 & 30.4 \\
GPT-5.6-sol & 76.9 & 28.8 & 60.6 & 53.6 & 23.2 & 44.6 \\
GPT-5.6-terra & 76.7 & 34.4 & 60.8 & \best{59.3} & 1.7 & \best{59.3} \\
Doubao-2.1 & 53.2 & 18.2 & 40.9 & 35.0 & 0.0 & 35.0 \\
Gemini-3.6 & 65.5 & 2.5 & \best{63.0} & 58.3 & 0.0 & 58.3 \\
GPT-5.6-luna & 49.0 & 17.3 & 34.2 & 50.0 & 1.7 & 48.3 \\
sonnet-5 & 9.3 & 1.5 & 8.2 & 15.0 & 0.0 & 15.0 \\
Qwen3.7-max & 29.1 & 0.5 & 28.6 & 45.0 & 0.0 & 45.0 \\
Qwen3.7-plus & 12.4 & 1.4 & 11.5 & 36.7 & 0.0 & 36.7 \\
MiniMax-M3 & 22.8 & 1.8 & 21.0 & 21.7 & 0.0 & 21.7 \\
Mimo-v2.5 & 32.4 & 0.0 & 32.9 & 1.7 & 0.0 & 1.7 \\
\hline
\end{tabular}
\endgroup
\end{center}
\end{table}

\subsection{Generalization and long-horizon diagnostics}

\begin{table}[H]
\caption{Single-arm controlled generalization success rates (\%).  Each entry
uses 20 held-out seeds and reuses the corresponding base-task demonstration
summary.}
\label{tab:single-arm-generalization}
\begin{center}
\begingroup
\scriptsize
\setlength{\tabcolsep}{3.2pt}
\begin{tabular}{@{}p{2.5in}rrrr@{}}
\hline
Task family & \shortstack{Qwen3.8\\max} & \shortstack{GPT-5.6\\sol} &
\shortstack{GPT-5.6\\terra} & \shortstack{GPT-5.6\\luna} \\
\hline
\nolinkurl{grasp_single_bottle}            &\best{85}&80&60&20\\
\nolinkurl{grasp_single_bottle_upright}   &40&35&50&\best{55}\\
\nolinkurl{grasp_single_cube}              &\best{80}&50&40&15\\
\nolinkurl{place_cube_in_bowl}             &\best{70}&40& 5&20\\
\nolinkurl{place_cube_on_cube}             &\best{80}&65& 5&10\\
\nolinkurl{place_single_bottle_upright}    &\best{55}&20&15&10\\
\nolinkurl{place_single_cube}              &\best{65}&45&20& 0\\
\hline
\textit{Macro-average} &\best{67.86}&47.86&27.86&18.57\\
\hline
\end{tabular}
\endgroup
\end{center}
\end{table}

Table~\ref{tab:generalization-capabilities} summarizes the same held-out
trajectories at the module level.  These equal-weight means are descriptive
profiles rather than an overall ranking.

\begin{table}[htbp]
\caption{Overall module-level diagnostic overview (\%) under controlled
held-out transfer.  SPU, RM, and ER equally average TL/AE/SR, MS/MP/FG, and
ED/OC/PF, respectively.}
\label{tab:generalization-capabilities}
\begin{center}
\begingroup
\small
\setlength{\tabcolsep}{5pt}
\begin{tabular}{lrrr}
\hline
Condition & SPU & RM & ER \\
\hline
Qwen3.8-max & 99.8 & \best{86.7} & 59.8 \\
GPT-5.6-sol & 99.3 & 76.4 & 59.9 \\
GPT-5.6-terra & \best{100.0} & 66.2 & \best{60.2} \\
GPT-5.6-luna & \best{100.0} & 54.3 & 34.4 \\
\hline
\end{tabular}
\endgroup
\end{center}
\end{table}

\subsection{Demonstration-summary source ablation}

Table~\ref{tab:summary-source-ablation} gives the subset breakdown for the
matched Qwen3.7-plus comparison summarized in
Section~\ref{sec:generalization}.

\begin{table}[htbp]
\caption{Effect of demonstration-summary source for Qwen3.7-plus.  Columns are
macro-averages over 11 single-arm tasks, three dual-arm tasks, and all 14
tasks, respectively.}
\label{tab:summary-source-ablation}
\begin{center}
\small
\begin{tabular}{lrrr}
\hline
Summary source & Single-arm & Dual-arm & All 14 \\
\hline
Qwen3.7-plus (model-specific) & \best{9.09} & \best{0.00} & \best{7.14} \\
Qwen3.8-max (shared)          & 6.36 & \best{0.00} & 5.00 \\
Shared $-$ model-specific     & $-2.73$ & 0.00 & $-2.14$ \\
\hline
\end{tabular}
\end{center}
\end{table}

Table~\ref{tab:summary-cross-eval-tasks} gives the per-task breakdown of the
cross-model shared-summary comparison summarized in
Section~\ref{sec:summary-source}.  Each entry is the success rate over the same
20 verified seeds used in the corresponding first base-evaluation run; only the
demonstration summary is replaced.  The own-summary block of each summary author
reproduces its base run exactly.

\begin{table}[htbp]
\caption{Per-task success rate (\%) under shared demonstration summaries, 20
seeds per cell.  Tasks are B = \texttt{grasp\_single\_bottle}, P =
\texttt{grasp\_single\_pen}, C = \texttt{place\_cube\_in\_bowl}.  \emph{Own} is
the model's own model-specific summary in its first base-evaluation run.}
\label{tab:summary-cross-eval-tasks}
\begin{center}
\small
\setlength{\tabcolsep}{2.6pt}
\begin{tabular}{@{}lrrrrrrrrrrrr@{}}
\hline
& \multicolumn{3}{c}{Own} & \multicolumn{3}{c}{On Qwen3.8-max} &
\multicolumn{3}{c}{On GPT-5.6-sol} & \multicolumn{3}{c}{On Opus-5} \\
\cline{2-4}\cline{5-7}\cline{8-10}\cline{11-13}
Evaluated model & B & P & C & B & P & C & B & P & C & B & P & C \\
\hline
Qwen3.8-max   & \best{95} & 80 & 85 & \best{95} & 80 & 85 & 90 & \best{80} & 85 & \best{95} & \best{85} & 85 \\
Opus-5        & 75 & \best{80} & \best{100} & 55 & \best{40} & \best{75} & 70 & 75 & \best{90} & 75 & 80 & \best{100} \\
GPT-5.6-sol   & 90 & 35 & 95 & 30 & 15 & 20 & \best{90} & 35 & \best{95} & 90 & 45 & 95 \\
GPT-5.6-terra & 55 & 55 & 15 & 20 & 5 & 5 & 65 & 30 & 30 & 60 & 65 & 40 \\
Gemini-3.6    & 40 & 30 & 0 & 15 & 15 & 0 & 60 & 30 & 15 & 50 & 35 & 20 \\
\hline
Mean          & 71.00 & 56.00 & 59.00 & 43.00 & 31.00 & 37.00 & 75.00 & 50.00 & 63.00 & 74.00 & 62.00 & 68.00 \\
\hline
\end{tabular}
\end{center}
\end{table}

\subsection{Locally deployed small-model results}

Table~\ref{tab:local-small-model-tasks} gives the task-level results summarized
in Table~\ref{tab:local-small-model-summary}.  Each task contains 20
trajectories.  Gemma-4-31B-it and the two Qwen3.6 variants use the Qwen3.8-max
demonstration summary, while Qwen3.8-27B uses a mixed-summary condition whose
summary source is also Qwen3.8-max.  Without summary guidance, Gemma-4-31B-it
and the two Qwen3.6 variants obtain $0.00\%$ on every task.

\begin{table}[htbp]
\caption{Summary-guided success for locally deployed models.  Entries are
success count / total trajectories followed by the success rate.  The
single-arm, dual-arm, and overall denominators are 220, 60, and 280,
respectively.}
\label{tab:local-small-model-summary}
\begin{center}
\begingroup
\small
\setlength{\tabcolsep}{3.5pt}
\begin{tabular}{llrrr}
\hline
Condition & Summary & Single-arm & Dual-arm & All 14 \\
\hline
Qwen3.8-27B      & Qwen3.8-max & \best{27/220 (12.27\%)} & 0/60 (0.00\%) & \best{27/280 (9.64\%)} \\
Gemma-4-31B-it   & Qwen3.8-max & 24/220 (10.91\%) & 0/60 (0.00\%) & 24/280 (8.57\%) \\
Qwen3.6-27B      & Qwen3.8-max & 3/220 (1.36\%) & 0/60 (0.00\%) & 3/280 (1.07\%) \\
Qwen3.6-35B-A3B  & Qwen3.8-max & 2/220 (0.91\%) & 0/60 (0.00\%) & 2/280 (0.71\%) \\
\hline
\end{tabular}
\endgroup
\end{center}
\end{table}

\begin{table}[H]
\caption{Per-task success rates (\%) for summary-guided locally deployed
models.  Each task contains 20 trajectories.  Bold blue marks the highest
nonzero result in each row.}
\label{tab:local-small-model-tasks}
\begin{center}
\begingroup
\scriptsize
\setlength{\tabcolsep}{3.2pt}
\renewcommand{\arraystretch}{0.94}
\begin{tabular}{@{}p{2.20in}rrrr@{}}
\hline
Task & \shortstack{Gemma-4\\31B-it} & \shortstack{Qwen3.6\\27B} &
\shortstack{Qwen3.6\\35B-A3B} & \shortstack{Qwen3.8\\27B} \\
\hline
\multicolumn{5}{@{}l}{\textit{Single-arm tasks}} \\
\nolinkurl{click_bell_right}                 & \best{10.00} & 0.00 & 0.00 & \best{10.00} \\
\nolinkurl{grasp_pen_leaning_cube}          & 5.00 & 0.00 & 0.00 & \best{10.00} \\
\nolinkurl{grasp_single_bottle}             & \best{30.00} & 0.00 & 5.00 & 15.00 \\
\nolinkurl{grasp_single_bottle_upright}     & 55.00 & 15.00 & 5.00 & \best{85.00} \\
\nolinkurl{grasp_single_cube}               & 5.00 & 0.00 & 0.00 & \best{10.00} \\
\nolinkurl{grasp_single_pen}                & \best{15.00} & 0.00 & 0.00 & 5.00 \\
\nolinkurl{place_cube_in_bowl}              & 0.00 & 0.00 & 0.00 & 0.00 \\
\nolinkurl{place_cube_on_cube}              & 0.00 & 0.00 & 0.00 & 0.00 \\
\nolinkurl{place_single_bottle_upright}     & 0.00 & 0.00 & 0.00 & 0.00 \\
\nolinkurl{place_single_cube}               & 0.00 & 0.00 & 0.00 & 0.00 \\
\nolinkurl{beat_block_hammer_right}         & 0.00 & 0.00 & 0.00 & 0.00 \\
\multicolumn{5}{@{}l}{\textit{Dual-arm tasks}} \\
\nolinkurl{place_shoe}                      & 0.00 & 0.00 & 0.00 & 0.00 \\
\nolinkurl{handover_horizontal_block}       & 0.00 & 0.00 & 0.00 & 0.00 \\
\nolinkurl{lift_pot}                        & 0.00 & 0.00 & 0.00 & 0.00 \\
\hline
\textit{Single-arm macro-average}           & 10.91 & 1.36 & 0.91 & \best{12.27} \\
\textit{Dual-arm macro-average}             & 0.00 & 0.00 & 0.00 & 0.00 \\
\textit{All-14 macro-average}               & 8.57 & 1.07 & 0.71 & \best{9.64} \\
\hline
\end{tabular}
\endgroup
\end{center}
\end{table}

\section{Scope and Limitations}
\label{app:scope-limitations}

\subsection{What the diagnostics and protocol comparisons identify}

The nine criteria are observational: they record where in a trajectory the
behavior visibly breaks, and they support the localization claims of
Section~\ref{sec:interaction-analysis}.  Attributing a breakpoint to perception,
planning, numerical grounding, or execution would require interventions on those
components, which this benchmark does not perform.  The same applies to the
protocol comparisons.  The active--passive result of
Section~\ref{sec:camera-protocol-ablation} shows, for the four conditions
tested, that removing camera control costs success even when passive input
supplies all five canonical views at each observation.  These fixed views form
a strict subset of the viewpoints reachable in active mode, which additionally
allows bounded incremental refinement.  The comparison leaves open whether the
cost comes from view relevance, cross-view integration, or the longer visual context, and it
says nothing about whether multiple simultaneous views are harmful in other
settings.  Similarly, the
cross-model summary comparison of Section~\ref{sec:summary-source} varies the
summary source over three tasks in a single run.  It therefore shows that the
source matters and that the ordering of the three sources is not the ordering of
their authors' task success, but the summaries differ in style as well as in
constraint explicitness, so it does not identify which property carries the
effect.

\subsection{Generalization scope}

The held-out variants are finite and controlled; they do not establish
open-set generalization across arbitrary objects, robots, cameras, or
real-world calibration.  The long-horizon result covers one compositional
family, and the local-model comparison covers four deployments under
summary-guided conditions.  Broader causal claims require additional object
families and deployment settings.

\subsection{Limitations: demonstration-summary conditioning}

The primary results are end-to-end condition comparisons: each model distills
its own demonstration summary and then uses it during control.  Outcomes
therefore combine summary quality, live-scene re-grounding, and action
generation rather than measuring summary-independent online control.  The
shared-summary comparison of Section~\ref{sec:summary-source} controls the
summary source across five models, but only over three tasks in a single run,
and the matched Qwen3.7-plus study changes it for one model over the full task
set.  Fully separating these factors requires no-summary and expert-written
summary controls across multiple models on the complete suite, which would also
distinguish constraint explicitness from other stylistic differences between
summaries.

\subsection{Limitations: modular VLM--VLA control}

The present benchmark evaluates a direct-control setting in which the VLM
produces metric robot and camera commands through a fixed, model-agnostic
harness.  We did not evaluate a modular architecture in which a VLM performs
high-level reasoning and instruction generation while a separately trained VLA
translates those instructions into low-level robot actions
\citep{zitkovich2023rt2,openx2024,ghosh2024octo,kim2024openvla,
li2024roboflamingo}. Consequently, the
reported results do not measure the benefits or failure modes introduced by a
VLM--VLA interface, a learned action policy, or a division of responsibility
between high-level planning and low-level control
\citep{shridhar2022cliport,shridhar2023peract,brohan2023rt1,jiang2023vima,
goyal2023rvt,chi2023diffusionpolicy}. Settling how such a
decomposition compares would require a common VLA controller, an explicit
instruction interface, and matched training and calibration protocols.

\section{Benchmark Specification Details}
\label{app:benchmark-specification}

VA-Bench evaluates whether a general-purpose MLLM can turn partial visual
evidence into spatially grounded robot behavior.  In the canonical protocol,
each episode couples active
observation, cross-view reasoning, numerical end-effector control, and
physics-simulated task completion.  The benchmark does not ask the model to
select an answer or a reference trajectory.  Instead, the model must determine
what evidence is missing, acquire an informative view, infer object--robot
geometry, and instantiate that inference as a sequence of metric actions.  The
camera interface deliberately isolates task-conditioned viewpoint selection
from physical camera-arm control, while the trajectory rubric reports
behavioral evidence rather than direct measurements of latent cognitive states.
The matched passive five-view comparison instead supplies five predefined RGB
views in each observation and exposes no camera-action command.  Robot actions,
seeds, execution, scoring, and context-budget settings remain the same.  Each
passive API request contains only the current five-image bundle; prior robot
actions, planner outcomes, and textual feedback remain in the retained history,
subject to the same context limit and compaction policy as the active condition.
Thus the comparison isolates the evidence-acquisition protocol rather than
robot control or history retention.

\subsection{Interactive task formulation}
\label{sec:app-task-formulation}

The following formulation describes the canonical active-view protocol.  An
evaluation instance is
\begin{equation}
    z=(\tau,\xi,g_{\tau,\xi},D_{\tau},B_{\tau},\phi_{\tau}),
    \label{eq:benchmark-instance}
\end{equation}
defined by a task family $\tau$, a scene seed $\xi$,
a natural-language goal $g_{\tau,\xi}$, an optional observation-only
demonstration $D_\tau$, an episode budget $B_\tau$, and a physics-based
task-success predicate $\phi_\tau$.  Before test-time control, the designated
summary model converts $D_\tau$ into a condition-specific textual summary
$s_\tau^{(c)}$; the primary condition uses the evaluated model itself, whereas
shared-summary controls use the named source model.  The simulator maintains an
unobserved physical state $x_t$.  At step $t$, the model receives

\begin{equation}
    o_t = \left(I_t, p_t, h_t, f_{t-1}\right),
\end{equation}

where $I_t$ is the current active-camera RGB image, $p_t$ is filtered robot
proprioception, $h_t$ is the interaction history, and $f_{t-1}$ contains the
previous action-validity and trajectory-planning outcomes.  At each decision,
the MLLM may call a non-environment inspection or demonstration tool, stop, or
choose one camera or robot action.  Let
$\mathcal{A}_{\mathrm{robot}}$ and $\mathcal{A}_{\mathrm{cam}}$ denote the
robot and camera action sets, respectively:

\begin{equation}
    a_t \sim M_\theta\!\left(g_{\tau,\xi},s_\tau^{(c)},o_{\leq t},a_{<t}\right),
    \qquad
    \begin{cases}
    x_{t+1}=\mathcal{T}_\tau(x_t,a_t), & a_t\in\mathcal{A}_{\mathrm{robot}},\\
    x_{t+1}=x_t, & a_t\in\mathcal{A}_{\mathrm{cam}}.
    \end{cases}
\end{equation}

and observes the resulting scene before choosing again.  Robot actions update
the physical state, whereas camera actions update only the observation pose.
Tool calls and stop decisions do not advance the environment; camera and robot
actions consume the same task-dependent budget.  Indexing dispatched
environment actions by $t$, an episode ends at
$t_{\mathrm{end}}\leq B_\tau$ when $\phi_\tau(x_{t_{\mathrm{end}}})$ becomes
true, the model stops, or the budget is exhausted.  The primary outcome is
therefore binary physics-simulated task success, rather than agreement with an
annotated answer or action sequence.

For the passive five-view comparison, the single actively controlled image
$I_t$ is replaced by a fixed-order bundle of five predefined RGB views, and
$\mathcal{A}_{\mathrm{cam}}$ is removed from the model's available action set.
The passive API sends only the current bundle; prior robot actions, planner
outcomes, and textual feedback remain in a history governed by the same context
limit and compaction policy as the canonical evaluation.  The other observation
fields, robot action space, action timing, seeds, executor, budget, success
predicate, and scoring are identical.  The model cannot select, reorder, or
request an additional view in this condition.

Every RGB frame in the observation payload is clean: model-facing images
contain no drawn keypoints, coordinate axes, masks, or target markers.  Proprioception
contains the controlled end-effector position and orientation, gripper
open/closed state, the physical finger midpoint (GC), and the current
gripper-local axes expressed in the world frame.  In dual-arm episodes, these
quantities are provided separately for both arms.  The model retains a
persistent history of prior robot actions, planner outcomes, and textual
feedback, bounded by the same context limit and compaction policy in both
conditions.  The passive API supplies only the current five-view bundle and
cannot obtain additional views.  The model is not given depth, point clouds,
segmentation,
object coordinates or poses, contact state, oracle grasp points, task
waypoints, or numerical camera calibration.  Consequently, the numerical robot
state anchors a metric reference frame by design; the benchmark does not test
recovery of the robot coordinate system itself.  The task-relevant object
geometry and its relation to that frame must still be inferred from RGB and
interaction.

\subsection{Task-level spatial demands}
\label{sec:app-spatial-demands}

At the task-design level, VA-Bench spans seven spatial--action demands that
become jointly observable only when perception is coupled to action.

\paragraph{Active evidence acquisition.}
The initial observation is not assumed to expose every task-relevant dimension.
The model must decide whether uncertainty concerns world-$XY$ position, height,
depth, orientation, clearance, or occlusion, and select a viewpoint that is
diagnostic for that uncertainty.  Because observation draws on the shared action
budget, the model must also stop looking and act once the accumulated evidence
is sufficient.

\paragraph{Cross-view consistency and online state tracking.}
Camera motion changes image coordinates but not the world-frame robot controls.
The model must reconcile observations across viewpoints, retain object identity
and spatial relations, and invalidate stale evidence after either the camera,
gripper, or object moves.  After execution, it must interpret the new image and
planner outcome relative to its previous action rather than restarting from an
independent frame.

\paragraph{Metric object--robot grounding.}
The model must translate qualitative visual relations into signed world-axis
translations and numerical distances.  This requires relating an object seen
in pixels to GC and end-effector coordinates, distinguishing camera-frame
appearance from world-frame motion, and estimating whether the remaining
lateral, vertical, and depth offsets are within grasp or placement tolerance.

\paragraph{Orientation, affordance, and contact geometry.}
Successful interaction depends on more than object centering.  The model must
infer long axes, surfaces, support relations, graspable regions, functional
parts, and approach corridors.  Examples include orienting the gripper closing
axis across a bottle or pen, grasping a hammer by its handle while aligning the
head with a block, pressing the top center of a bell, and preserving an upright
container during placement.

\paragraph{Goal-directed metric action sequencing and outcome-conditioned revision.}
The model must compose approach, grasp, lift, transport, insertion, release,
and retreat phases using only incremental metric commands.  Low-level inverse
kinematics, collision checking, and trajectory generation remain delegated to
the fixed execution layer.
Re-observation after each action enables online correction when a proposed
translation is rejected, an object moves unexpectedly, a grasp misses, or a
placement is not yet stable.  The trajectory-level diagnostics measure the
resulting recovery behavior at the level of observable actions
(Appendix~\ref{app:scope-limitations}).

\paragraph{Embodiment and bimanual coordination.}
Tasks with two available arms require reasoning about reachability, arm--object
assignment, shared-workspace clearance, and temporal dependencies between
roles.  Handover tasks require complementary grasp regions and ordered release,
whereas pot lifting requires mirrored handle geometry and synchronized motion.

\paragraph{Demonstration-conditioned spatial re-grounding.}
The demonstration communicates visible task procedure, not the live scene's
metric solution.  The model must extract transferable constraints and
re-ground them under a new seed, object pose, arm assignment, or held-out
geometry.  This separates recognizing a demonstrated procedure from executing
it in the current spatial configuration.

These seven demands are design dimensions that describe what the task suite
requires; they are not seven additional scores or independently identified
latent variables.  Section~\ref{sec:app-diagnostic-scoring} instead operationalizes
executed behavior through nine scored trajectory-level diagnostics in three
modules.  The two levels are intentionally not one-to-one: for example,
bimanual coordination is a task condition and demonstration-conditioned
re-grounding is evaluated through the transfer protocol.  End-to-end outcomes
also depend on numerical action generation, the fixed motion planner, and
simulated dynamics.

\subsection{Task-family coverage}
\label{sec:app-task-suite}

VA-Bench comprises 14 base task families: eleven expose a single controllable
arm, while three expose two arms and require the model to reason explicitly
about their use.  In
\nolinkurl{place_shoe}, the model must select and control the arm on the
object's reachable side; \nolinkurl{handover_horizontal_block} and
\nolinkurl{lift_pot} additionally require coordinated use of both arms.  We
therefore classify all three as dual-arm tasks, regardless of whether both arms
move simultaneously.

In terms of task demands, the single-arm suite tests target pose and long-axis
inference, object--robot alignment, grasp-region and approach-direction
selection, and placement constraints such as containment, stacking, pose
preservation, and release height.  Bell pressing and hammering further test
functional-part localization, contact direction, and tool geometry.  The
dual-arm suite requires reachability-aware arm assignment, ordered handover,
or synchronized motion.  The separately evaluated five-object task tests
long-horizon state maintenance, repeated spatial re-grounding, and composition
of atomic skills.
These demands describe task coverage rather than additional independent scores.

\paragraph{Instances and controlled splits.}
A scene seed selects task-specific object positions, orientations, and, where
configured, asset model IDs.  Resulting layouts or arm assignments follow from
these task fields rather than from a separate global randomizer.  The base
benchmark uses 20 retained seeds per task under the verification protocol in
Appendix~\ref{sec:app-seed-validation}.  The primary terminal-success
evaluation repeats this fixed suite in three independent runs per model;
probe-only simulator state and actions
are never exposed to the evaluated model.  Seven base families additionally
have paired held-out variants: horizontal and upright bottle grasping, cube
grasping, cube-in-bowl placement, cube stacking, upright-bottle placement, and
cube placement.  Each variant reuses its base-task demonstration while changing
controlled object geometry, container or box-like instance, or spatial layout.
The five-object basket task is a separate compositional track conditioned on
four atomic-skill videos rather than a full-task demonstration.

\subsection{Seed construction and physical verification}
\label{sec:app-seed-validation}

\paragraph{Deterministic candidate instantiation.}
Before environment initialization, each integer seed fixes the NumPy and Torch
random states.  Evaluation uses \texttt{demo\_clean}: global background,
lighting, camera, table-height, and clutter randomization are disabled, leaving
only the task-coded object position, orientation, and model-ID fields listed
below.  Consequently, a fixed task implementation and seed reproduce the same
scene.  The task-dependent \texttt{candidate\_seeds} field is an ordered pool,
not a declaration that every listed candidate was instantiated.  Configured
pools use task-dependent integer ranges 1000--1099, 1000--1119, 1000--1199,
1000--1299, or 1000--5000, but these bounds describe the available search order
rather than the number of executed probes.  Candidates are probed in order,
and collection stops as soon as 20 accepted seeds have been obtained for a
task.

\paragraph{Orientation-based settling and acceptance gate.}
Each instantiated scene is reset, advanced for 2,000 physics steps, and then
observed for 500 additional steps.  During the final 200 observation steps,
explicit object-orientation drift must remain within approximately
$3^\circ$.  This is an orientation-based settling check; the current selector
does not impose a separate translational-drift threshold.  Initial and
post-probe object poses must also be finite.  A seed receives
\texttt{ready=true} only when its privileged probe, the planner
result, terminal checker, gripper state, and task-specific physical
outcomes all pass.  A probe run that ends before producing a physical outcome
is recorded separately and does not by itself classify the scene as physically
infeasible.

\paragraph{Verification scope.}
All 280 retained manifest entries have a matching \texttt{ready=true} probe
under a single verification mode: a complete \texttt{task.play\_once()}
execution with the full planner chain, \texttt{planner\_success}, the terminal
checker, gripper state, and the task-specific physical outcome all passing.
Every retained seed therefore carries a verified expert execution from the
robot home state through the terminal predicate, so each base-task instance is
certified physically solvable end to end rather than merely reachable at the
grasp endpoint.  Across
all tasks, 318 candidates were actually instantiated and probed before the
per-task stopping rule was met, and 280 were retained.  The resulting 88.1\%
is a conditional yield among candidates that were run, not an acceptance rate
over the complete task-dependent candidate pools (Table
\ref{tab:seed-probe-yield}).

\begin{table}[htbp]
\caption{Conditional seed yield among candidates actually instantiated and
probed.  Selection visits each ordered pool only until 20 accepted seeds are
collected, so unvisited pool entries are not part of the denominator.}
\label{tab:seed-probe-yield}
\centering
\begingroup
\small
\setlength{\tabcolsep}{4pt}
\begin{tabular}{@{}p{3.18in}rrr@{}}
\hline
Task or group & Probed & Retained & Yield (\%) \\
\hline
Ten tasks whose first 20 probes passed & 200 & 200 & 100.0 \\
\nolinkurl{place_cube_on_cube} & 22 & 20 & 90.9 \\
\nolinkurl{place_single_cube} & 22 & 20 & 90.9 \\
\nolinkurl{beat_block_hammer_right} & 24 & 20 & 83.3 \\
\nolinkurl{lift_pot} & 50 & 20 & 40.0 \\
\hline
All 14 tasks & 318 & 280 & 88.1 \\
\hline
\end{tabular}
\endgroup
\end{table}

\paragraph{Task randomization and terminal predicates.}
Tables~\ref{tab:seed-spec-grasp}--\ref{tab:seed-spec-dual} report the task-coded
scene fields and terminal checker conditions used by the base suite.  Position
ranges are in world-frame meters; tolerances are stated explicitly in
millimeters.  ``Lifted'' is measured relative to the corresponding initial
object height.  TCP denotes the tool-center point.

\begin{table}[p]
\caption{Seed randomization and terminal predicates for direct-grasp tasks.}
\label{tab:seed-spec-grasp}
\centering
\begingroup
\scriptsize
\renewcommand{\arraystretch}{0.94}
\setlength{\tabcolsep}{2.5pt}
\begin{tabular}{@{}>{\raggedright\arraybackslash}p{1.34in}>{\raggedright\arraybackslash}p{1.90in}>{\raggedright\arraybackslash}p{2.01in}@{}}
\hline
Task & Task-coded scene fields & Terminal success predicate \\
\hline
\nolinkurl{grasp_single_bottle} & $x\in[0.20,0.28]$,
$y\in[-0.16,-0.04]$; yaw $\in[-22.5^\circ,22.5^\circ]$ & Target lifted
$>50$~mm; right gripper closed \\
\nolinkurl{grasp_single_bottle_upright} & $x\in[0.20,0.28]$,
$y\in[-0.16,-0.04]$; fixed upright orientation & Target lifted $>50$~mm;
right gripper closed \\
\nolinkurl{grasp_single_cube} & $x\in[0.18,0.26]$,
$y\in[-0.16,-0.06]$; yaw $\in[-45^\circ,45^\circ]$ & Target lifted
$>50$~mm; right gripper closed \\
\nolinkurl{grasp_single_pen} & $x\in[0.18,0.26]$,
$y\in[-0.20,-0.10]$; yaw $\in[-30^\circ,30^\circ]$ & Target lifted
$>50$~mm; right gripper closed \\
\nolinkurl{grasp_pen_leaning_cube} & Group $x\in[0.16,0.22]$,
$y\in[-0.24,-0.20]$; yaw $\in[-15^\circ,15^\circ]$; fixed $30^\circ$
lean & Pen lifted $>50$~mm; right gripper closed \\
\hline
\end{tabular}
\endgroup
\end{table}

\begin{table}[p]
\caption{Seed randomization and terminal predicates for placement and stacking
tasks.}
\label{tab:seed-spec-place}
\centering
\begingroup
\scriptsize
\renewcommand{\arraystretch}{0.94}
\setlength{\tabcolsep}{2.5pt}
\begin{tabular}{@{}>{\raggedright\arraybackslash}p{1.34in}>{\raggedright\arraybackslash}p{1.90in}>{\raggedright\arraybackslash}p{2.01in}@{}}
\hline
Task & Task-coded scene fields & Terminal success predicate \\
\hline
\nolinkurl{place_cube_in_bowl} & Bowl $x\in[0.10,0.14]$,
$y\in[0,0.04]$; cube $x\in[0.22,0.27]$, $y\in[-0.17,-0.10]$; cube yaw
$\in[-45^\circ,45^\circ]$ & Cube previously lifted $\geq50$~mm; center
distance $<55$~mm; relative height 15--95~mm; bowl displacement $<5$~mm;
contact; gripper open \\
\nolinkurl{place_cube_on_cube} & Red cube $x\in[0.20,0.26]$,
$y\in[-0.14,-0.06]$; base cube $x\in[0.10,0.14]$, $y\in[0,0.04]$; yaw
$\in[-45^\circ,45^\circ]$ & Red cube previously lifted $\geq50$~mm;
top-surface overlap $\geq2$~mm; height error $\leq12$~mm; base displacement
$\leq12$~mm; contact; gripper open \\
\nolinkurl{place_single_cube} & Cube as in
\nolinkurl{grasp_single_cube}; target $x\in[0.10,0.14]$, $y\in[0,0.04]$ &
Cube previously lifted $\geq50$~mm; $x/y$ error each $\leq35$~mm; height error
$\leq35$~mm; target displacement $\leq2$~mm; contact; gripper open \\
\nolinkurl{place_single_bottle_upright} & Bottle as in the upright-grasp
task; target $x\in[0.03,0.09]$, $y\in[0.07,0.13]$ & Bottle previously lifted
$\geq50$~mm; $x/y$ error each $\leq40$~mm; height error $\leq35$~mm; target
displacement $\leq2$~mm; contact; gripper open \\
\hline
\end{tabular}
\endgroup
\end{table}

\begin{table}[p]
\caption{Seed randomization and terminal predicates for functional-contact and
tool-use tasks.}
\label{tab:seed-spec-functional}
\centering
\begingroup
\scriptsize
\renewcommand{\arraystretch}{0.94}
\setlength{\tabcolsep}{2.5pt}
\begin{tabular}{@{}>{\raggedright\arraybackslash}p{1.34in}>{\raggedright\arraybackslash}p{1.90in}>{\raggedright\arraybackslash}p{2.01in}@{}}
\hline
Task & Task-coded scene fields & Terminal success predicate \\
\hline
\nolinkurl{click_bell_right} & $x\in[0.12,0.26]$,
$y\in[-0.18,-0.04]$; two bell model IDs & Gripper closed; contact point
within 25~mm of the bell functional point on each of $x$ and $y$, and within
30~mm on $z$ \\
\nolinkurl{beat_block_hammer_right} & Hammer $x\in[0.02,0.06]$,
$y\in[-0.10,-0.04]$; block $x\in[0.14,0.22]$, $y\in[0.06,0.13]$; yaw
$\in[-0.5,0.5]$~rad & Hammer lifted with contact by $>40$~mm; functional-point
$x/y$ error each $<20$~mm; physical hammer--block contact; gripper closed \\
\hline
\end{tabular}
\endgroup
\end{table}

\begin{table}[p]
\caption{Seed randomization and terminal predicates for dual-arm tasks.}
\label{tab:seed-spec-dual}
\centering
\begingroup
\scriptsize
\renewcommand{\arraystretch}{0.94}
\setlength{\tabcolsep}{2.5pt}
\begin{tabular}{@{}>{\raggedright\arraybackslash}p{1.34in}>{\raggedright\arraybackslash}p{1.90in}>{\raggedright\arraybackslash}p{2.01in}@{}}
\hline
Task & Task-coded scene fields & Terminal success predicate \\
\hline
\nolinkurl{place_shoe} & Shoe $x\in[-0.25,0.25]$,
$y\in[-0.10,0.05]$; ten model IDs; samples with horizontal radius $<0.15$~m
excluded & $x$ error $<50$~mm; $y$ error $<20$~mm; absolute error of each
quaternion component $<0.07$; both grippers open \\
\nolinkurl{handover_horizontal_block} &
$x\in[-0.20,-0.15]\cup[0.15,0.20]$, $y\in[-0.05,0]$ & Required handover state
chain completed after vertical presentation; giver open and receiver closed;
$z>0.92$~m; object on receiver side; long axis within $20^\circ$ of vertical \\
\nolinkurl{lift_pot} & $x,y\in[-0.05,0.05]$; yaw
$\in[-22.5^\circ,22.5^\circ]$; two pot model IDs & Pot $z>0.82$~m; each TCP
within 30~mm of its corresponding handle; upright dot product $>0.8$ \\
\hline
\end{tabular}
\endgroup
\end{table}

The shared target-placement predicate is implemented in
\nolinkurl{envs/_single_right_arm_place_target_task.py}; the remaining terminal
predicates reside in their corresponding task environments.  Deterministic
initialization and the \texttt{demo\_clean} policy are implemented in
\nolinkurl{envs/_base_task.py} and
\nolinkurl{task_config/demo_clean.yml}, while ordered candidate screening and
both probe modes are implemented in
\nolinkurl{codex-robotwin-agent/scripts/select_safe_eval_seeds.py}.

\subsection{Metric robot action space}
\label{sec:app-robot-action-space}

The environment action space is

\begin{equation}
    \mathcal{A}=\mathcal{A}_{\mathrm{cam}}\cup
    \mathcal{A}_{\mathrm{move}}\cup
    \mathcal{A}_{\mathrm{rot}}\cup
    \mathcal{A}_{\mathrm{grip}}\cup
    \mathcal{A}_{\mathrm{dual}}.
\end{equation}

Table~\ref{tab:app-robot-actions} gives the exact robot-command parameterization.
The interface exposes numerical Cartesian target changes rather than the
legacy small/medium/large motion labels or low-level motor commands.

\begin{table}[ht]
\caption{Robot actions exposed to the MLLM.  In dual-arm mode, every
single-gripper numerical command must name the controlled arm.}
\label{tab:app-robot-actions}
\small
\begin{center}
\begin{tabular}{@{}>{\raggedright\arraybackslash}p{1.02in}
                    >{\raggedright\arraybackslash}p{1.46in}
                    >{\raggedright\arraybackslash}p{2.65in}@{}}
\hline
Action family & Parameters & Semantics \\
\hline
World translation
& $u\!\in\!\{x,y,z\}$, sign $\sigma\!\in\!\{-,+\}$, $d\!\in\![1,100]$ mm
& Move the selected end effector by $\sigma d$ along a fixed world axis; the target is bounded by the robot workspace. \\

Local rotation
& $r\!\in\!\{r_x,r_y,r_z\}$, sign $\sigma$, $\alpha\!\in\![1,90]^\circ$
& Rotate about the selected current gripper-local axis using the right-hand rule; compensated waypoints keep GC approximately fixed. \\

Gripper aperture
& \texttt{open}, \texttt{close}
& Command the selected physical gripper without prescribing an object or contact state. \\

Synchronized dual translation
& Independent $(u,\sigma,d)$ for left and right
& Plan and execute both world-axis targets in one RoboTwin move call. \\

Synchronized dual aperture
& \shortstack[l]{\texttt{dual\_gripper.open}\\\texttt{dual\_gripper.close}}
& Open or close both grippers in the same environment step. \\
\hline
\end{tabular}
\end{center}
\end{table}

These action families map to \texttt{gripper.move\_world},
\texttt{gripper.rotate\_local}, and gripper aperture commands.  Dual-arm
episodes require \texttt{arm=left|right} for single-gripper commands and add
the synchronized \texttt{dual\_gripper} actions listed above.  A fixed
model-agnostic IK and trajectory layer attempts each metric target and returns
binary planning feedback; it does not select the object, grasp region, arm
role, or task waypoint.  Section~\ref{sec:agent-harness} defines
this execution boundary.

\subsection{Bounded active camera action space}
\label{sec:app-camera-action-space}

The main evaluation exposes the camera actions in
Table~\ref{tab:app-camera-actions}.  Their semantic views are defined relative to GC,
not to a privileged target-object location.  Let $c$ be the selected arm's
finger midpoint (or the midpoint of both arms), and let the sign of the lateral
offset follow the selected arm side.  Each semantic-view action moves the
camera toward the listed nominal offset and orients it toward $c$; the final
pose is clamped to the camera workspace.

\begin{table}[t]
\caption{Camera actions in the active-view protocol.  Semantic viewpoints are
gripper-centered; incremental translations and rotations are expressed in the
current camera frame unless stated otherwise.}
\label{tab:app-camera-actions}
\small
\begin{center}
\begin{tabular}{@{}p{1.30in}p{1.64in}p{2.19in}@{}}
\hline
Action & Motion & Primary geometric evidence \\
\hline
\texttt{view\_topdown}
& $c+(0,0,0.68)$ m
& World-$XY$ displacement, planar yaw, target coverage \\

\texttt{view\_side}
& $c+(\pm0.58,-0.02,0.06)$ m
& Height, table clearance, vertical overlap and insertion \\

\texttt{view\_front\_side\_45}
& $c+(\pm0.44,-0.44,0.18)$ m
& Break front/side occlusion and horizontal projection ambiguity \\

\texttt{view\_side\_top\_45}
& $c+(\pm0.48,-0.02,0.48)$ m
& Joint evidence for lateral alignment, height, and finger placement \\

\texttt{view\_oblique\_45}
& $c+(\pm0.42,-0.42,0.42)$ m
& Overall 3D configuration and selection of the next diagnostic view \\

\shortstack[l]{\texttt{move\_left/right}\\\texttt{move\_up/down}}
& Translate 5 cm along camera left/right/up/down
& Local de-occlusion and small framing changes \\

\texttt{zoom\_in/out}
& Translate 8 cm along camera forward/backward
& Inspect or widen the local interaction region \\

\texttt{yaw\_left/right}
& Rotate $10^\circ$ around world $Z$
& Horizontal parallax and side exposure \\

\texttt{pitch\_up/down}
& Rotate $10^\circ$ around camera-left axis
& Vertical parallax and top/side exposure \\
\hline
\end{tabular}
\end{center}
\end{table}

In single-arm tasks, the five semantic views target the active gripper.  In
dual-arm tasks, generic views target the two-arm midpoint, while
\texttt{view\_left\_*}, \texttt{view\_right\_*}, and
\texttt{view\_both\_*} explicitly select the observation target.  Camera poses
are clamped to a bounded workspace.  The action changes only the observation
viewpoint and never changes the world-frame meaning of robot actions.

We keep camera control discrete because continuous parameters would add
control burden without making evidence acquisition more difficult.
This interface deliberately evaluates task-conditioned viewpoint selection and
bounded camera adjustment, not physical camera-arm motion.  Semantic views and
incremental camera actions directly update the simulated camera pose; camera
reachability, collision, continuous trajectory cost, and calibration error are
outside the current scope.  Each camera action still consumes an environment
step, and the agent instruction limits camera-only chains to at most five
consecutive actions.  Table~\ref{tab:app-camera-actions} is the complete set of
camera commands available in every reported active-view run; the passive
control condition of Appendix~\ref{sec:app-five-view-protocol} exposes none of
them.

\subsection{Passive five-view observation protocol}
\label{sec:app-five-view-protocol}

The matched Qwen3.8-max comparison supplies five clean, predefined RGB images at
each observation, in place of the single actively acquired image used by the
canonical condition.  The passive condition exposes no camera-action command.
The textual observation block appears first, followed by image blocks in the
same fixed order \texttt{topdown}, \texttt{side},
\texttt{front\_side\_45}, \texttt{side\_top\_45}, and
\texttt{oblique\_45}.  The local session record stores these view labels, but
the evaluated OpenAI-compatible API request includes no additional textual
label beside an image; the fixed block order is therefore the model-visible cue
to view identity.  The observer uses a $55^\circ$ vertical field of view, a
0.05-m near plane, and a 100-m far plane.  The images contain no end-effector or
target marker, segmentation mask, depth map, or debugging overlay.  These
camera parameters define the renderer but are not sent to the model as
numerical calibration.

The passive five-view bundle is supplied after reset and robot actions at the
same control points where the canonical protocol supplies its current image.
The active model may then select further views through camera actions, whereas
the passive model has no such command.  Its API observation contains only the
current five-image bundle; prior robot actions, planner outcomes, and textual
feedback remain under the same bounded-history and compaction policy as the
canonical evaluation.  Robot-action semantics and environment transitions are
also matched.  The comparison therefore contrasts model-directed active
evidence acquisition with passively supplied multi-view evidence.

\subsection{Demonstration-conditioned evaluation protocol}
\label{sec:app-demonstration-protocol}

For each standard task, the main protocol supplies one RGB-only video of a
successful expert execution.  Before evaluation, the evaluated MLLM retrieves
multiple sampled frames and writes a structured textual summary of visible
constraints, including grasp region, finger placement, approach, posture,
arm assignment, coordination order, and completion conditions.  The video
contains visible robot and object motion, and therefore conveys procedural
information, but exposes no machine-readable expert action, EEPose, joint
state, object coordinate, contact label, or semantic phase.  The summary is
task-level experience; it does not contain the metric solution for a live
seed.  This task-level procedure video is distinct from the per-seed acceptance
probes in Appendix~\ref{sec:app-seed-validation}; those probe executions and
their recorded trajectories remain evaluator-only and are never shown to the
evaluated model.  Demonstration interpretation and procedural abstraction are
therefore part of the primary end-to-end construct, rather than fixed
preprocessing.

For the primary model-specific-summary condition, this sequence is
\begin{equation}
    D_{\tau}\xrightarrow[\text{multiple distinct RGB frames}]{\text{model inspection}}
    s_{m,\tau}\xrightarrow[\text{new seed}]{\text{re-grounding}}
    \{a_t,o_t\}_{t=0}^{t_{\mathrm{end}}}.
    \label{eq:demo-regrounding}
\end{equation}
At test time, the model receives the task instruction, generic safety and
manipulation constraints, any task-specific semantic completion constraints,
and its demonstration-derived summary.  It must nevertheless reacquire the
current scene, choose views, assign arms where applicable, and generate every
numerical parameter.  No oracle trajectory, target waypoint, target-object
state, or task phase is supplied.  The model-specific-summary condition is the
primary end-to-end protocol.  We additionally report a shared-summary condition
as a controlled diagnostic that fixes the textual procedural context across
models and therefore partially isolates downstream re-grounding and execution.
Held-out variants reuse
the base-task summary without a new video,
while the composite basket task uses four independent atomic-skill videos and
no full-task demonstration.

\subsection{Trajectory-based capability diagnostics}
\label{sec:app-diagnostic-scoring}

Terminal task success is deliberately strict, but it does not reveal whether a
model failed to find the target, misunderstood the robot--object geometry,
selected an unsuitable motion, or was unable to recover.  We therefore augment
the environment outcome with a trajectory-level diagnostic rubric.  The rubric
decomposes task competence into three first-level modules---\emph{spatial
perception and understanding}, \emph{robot manipulation}, and \emph{error
recovery}---with three separately scored trajectory-level behavioral criteria
in each module.

\begin{figure}[H]
    \centering
    \includegraphics[width=\linewidth]{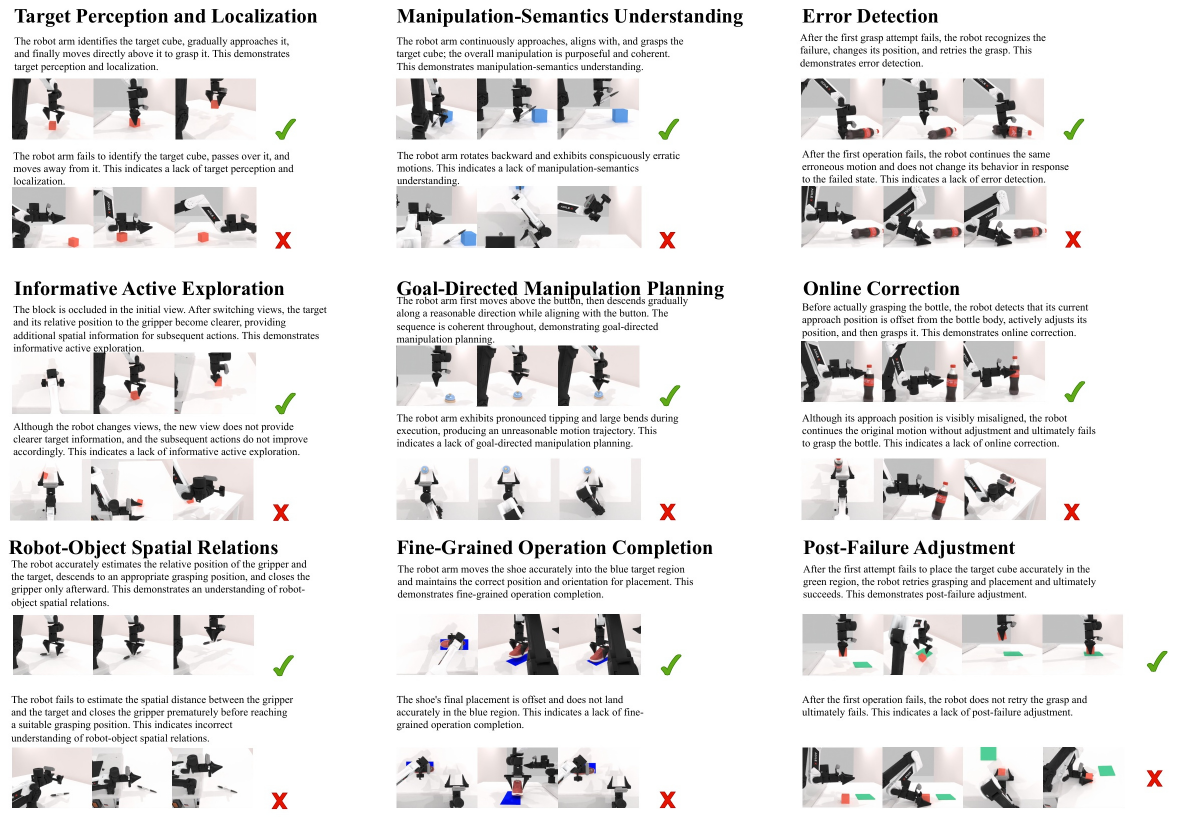}
    \caption{Annotation reference for the nine trajectory-level
    capability criteria.  In the eight trajectory panels, the upper row is a representative
    positive trajectory (green check) and the lower row is a representative
    negative trajectory (red cross).  The FG panel gives its pre-contact
    analysis rule; final placement alone does not establish this behavior.
    The examples illustrate observable evidence, not hidden model states.  For error
    detection (ED), online correction (OC), and post-failure adjustment (PF),
    the examples contain an observable error; error-free episodes remain NA.}
    \label{fig:annotation-rubric}
\end{figure}

\paragraph{Evidence and annotation unit.}
The annotation unit is one complete episode.  Each complete trajectory is
independently reviewed by three annotators using the same episode record and
rubric.  Annotation proceeds in two stages: the three annotators first judge
criterion applicability, and the majority decision fixes the episode-level
gate; for every activated criterion, all three then record a binary behavioral
judgment whose majority vote is the reported episode label.  Annotators inspect
the task instruction, the sequence of
model-facing RGB observations, camera and robot actions, resulting scene
changes, action validity, planner outcomes, and the terminal task state.  The
three independent judgments are retained in the annotation record, and the
aggregation equations below operate on the episode-level labels used for
reporting.  Detailed positive and negative evidence rules
and applicability conditions are provided in this appendix.  Figure~\ref{fig:annotation-rubric}
gives a visual reference for the decision
standard, with positive and negative trajectory examples for eight criteria
and a scoring guide for FG.  A
model's textual rationale can help identify its stated intent,
but does not by itself establish a capability: the executed trajectory must
contain corresponding behavioral evidence.  Conversely, a single planner
rejection is not automatically a planning failure, because a geometrically
reasonable target may still be infeasible for the fixed execution layer.  The
judgment instead considers whether the model subsequently changes its estimate
or repeatedly issues actions that remain inconsistent with the visible task.

For annotator $a\in\{1,2,3\}$, let $g_e^{(a)}\in\{0,1\}$ indicate whether
episode $e$ contains an observable execution error.  The shared recovery gate
is the majority vote
\begin{equation}
    g_e=\mathbb{1}\!\left[\sum_{a=1}^{3}g_e^{(a)}\geq 2\right].
    \label{eq:error-applicability-vote}
\end{equation}
If $g_e=0$, ED, OC, and PF are all recorded as $\bot$ (reported as NA) and the
episode is excluded from all three denominators.  If $g_e=1$, all three
criteria are activated.  For each activated criterion $k$, annotator $a$
records $r_{e,k}^{(a)}\in\{0,1\}$, where one denotes positive behavioral
evidence and zero denotes that the behavior was not demonstrated.  The
episode-level label used for reporting is
\begin{equation}
    r_{e,k}=\mathbb{1}\!\left[\sum_{a=1}^{3}r_{e,k}^{(a)}\geq 2\right].
    \label{eq:criterion-majority-vote}
\end{equation}
The same majority-vote rule is used for the binary labels of the other
criteria.  AE carries an analogous assessable-ambiguity gate, but in the
annotated runs every episode was judged to contain a task-relevant ambiguity
under single-view observation, so the gate never fired and AE is scored over all
annotated episodes; TL, SR, MS, MP, and FG are applicable by construction.
Thus, an episode is not penalized for having no error, but an applicable
capability is scored zero when at least two annotators find no corresponding
behavioral evidence.

\subsubsection{Annotation reliability}
\label{sec:app-annotation-reliability}

Because the nine criteria are human-annotated, we report their agreement
directly.  The annotation record retains, for every activated criterion of every
annotated episode, how many of the three annotators recorded positive evidence.
Across the 3,360 annotated base episodes of the 12 primary conditions, this
gives $29{,}232$ applicable episode--criterion items and $87{,}696$ individual
binary judgments.  Table~\ref{tab:annotation-agreement}
summarizes them.  Terminal task success is determined solely by the environment
checker and is never annotated, so annotation reliability bounds the
interpretation of the nine diagnostics and of subtask progress, not of any
success rate reported in this paper.

All $1{,}259$ disagreements are $2$--$1$ splits, and every reported label
matches the majority vote.  Overall unanimity is $95.69\%$, with Fleiss
$\kappa=0.942$ \citep{fleiss1971agreement}.

For criteria whose positive rate is far from balanced, $\kappa$ understates
agreement, because a skewed marginal inflates the chance-agreement term.  TL is
the clearest case: annotators agree unanimously on $97.41\%$ of episodes, but
the near-ceiling positive rate yields $\kappa=0.723$.  We also report Gwet
AC1, which is less sensitive to prevalence, and interpret the two together.
FG has the lowest unanimity ($87.38\%$, $\kappa=0.797$) and accounts for
$424$ of the split votes \citep{gwet2008agreement}.  Its individual scores warrant particular care when
comparing voting rules.

\begin{table}[t]
\centering
\small
\caption{Inter-annotator agreement over the nine trajectory criteria.  $n$ is the
number of applicable episode--criterion items (three judgments each); ED, OC, and PF are
scored only on episodes containing an observable error, hence their smaller $n$.
Unanimity is the fraction of items on which all three annotators agree; every
remaining judgment is a $2$--$1$ split.  Positive rate is the fraction of
majority-vote labels equal to one, and drives the difference between $\kappa$ and
AC1 for near-ceiling criteria.  Statistics cover the annotated base run of the 12
primary conditions; the held-out-transfer and long-horizon annotations are
excluded here and contained no split votes.}
\label{tab:annotation-agreement}
\setlength{\tabcolsep}{5pt}
\begin{tabular}{@{}llrrrrr@{}}
\hline
Module & Criterion & $n$ & Unanimity & Pos.\ rate & Fleiss $\kappa$ & Gwet AC1 \\
\hline
SPU & TL & 3{,}360 & 97.41\% & 0.974 & 0.723 & 0.982 \\
SPU & AE & 3{,}360 & 92.77\% & 0.376 & 0.898 & 0.908 \\
SPU & SR & 3{,}360 & 94.70\% & 0.789 & 0.892 & 0.948 \\
\hline
RM & MS & 3{,}360 & 99.88\% & 0.988 & 0.968 & 0.999 \\
RM & MP & 3{,}360 & 97.08\% & 0.565 & 0.960 & 0.962 \\
RM & FG & 3{,}360 & 87.38\% & 0.321 & 0.797 & 0.856 \\
\hline
ER & ED & 3{,}024 & 96.40\% & 0.447 & 0.951 & 0.953 \\
ER & OC & 3{,}024 & 99.40\% & 0.124 & 0.982 & 0.995 \\
ER & PF & 3{,}024 & 96.76\% & 0.369 & 0.953 & 0.960 \\
\hline
\multicolumn{2}{@{}l}{All criteria} & 29{,}232 & 95.69\% & 0.559 & 0.942 & 0.943 \\
\multicolumn{2}{@{}l}{All except FG} & 25{,}872 & 96.77\% & 0.589 & 0.956 & 0.958 \\
\hline
\end{tabular}
\end{table}

\paragraph{Robustness to the aggregation rule.}
Majority voting is one of several defensible ways to collapse three judgments.
We recomputed all nine criteria for all 12 primary conditions under two alternatives: a
strict rule that credits a criterion only when all three annotators are positive,
and a permissive rule that credits it when any annotator is positive.  Averaged
over the nine criteria, the largest within-condition range across the three
rules is $6.25$ points.  Relative to majority voting, the strict rule decreases
this mean by at most $3.93$ points and the permissive rule increases it by at
most $2.33$ points.  The induced orderings are similar (Spearman
$\rho\ge0.979$ and Kendall $\tau\ge0.909$ for every pairwise comparison).  The
three strongest conditions under majority voting remain the three strongest under
both alternatives.  Individual criteria are more sensitive: GPT-5.6-sol leads
FG under unanimity ($62.5\%$), whereas Opus-5 leads under majority voting
($77.9\%$).  Thus stability of the nine-criterion means does not establish
stability of every individual score or ordering.

\paragraph{Annotations are not restatements of the outcome.}
A concern for any trajectory-level rubric is that annotators may simply infer the
labels from whether the episode succeeded.  The record contradicts this.  On the
$1{,}680$ annotated episodes for which the per-episode file also records the
checker verdict, FG agrees with terminal success on $82.08\%$ of episodes.
There are $116$ episodes with FG$=0$ and checker success, and $185$ with FG$=1$
and checker failure.  FG measures pre-contact analysis, so a positive label
does not require completion, and success alone does not establish such
analysis.  These counts do not identify the cause of individual mismatches.
TL and MS behave in the
opposite way: they agree with terminal success on $31.1\%$ and $30.7\%$ of the same
episodes, because more than $1{,}100$ episodes are labeled positive on target
localization and manipulation semantics while still failing, and \emph{no}
episode is labeled negative on either criterion while succeeding.  Annotators
that copied the outcome could not produce this pattern.  The same asymmetry is
itself a finding: the visual and semantic prerequisites are usually satisfied,
and the failures accumulate downstream.

\paragraph{What the record does not support.}
The retained counts are per-episode tallies rather than per-annotator labels, so
statistics that require annotator identity---pairwise Cohen $\kappa$, per-rater
bias estimates, and Cochran's $Q$---are not recoverable and we do not report
them.  The concordance figures above are computed on the subset of annotation
files that carry a machine-readable success column ($1{,}680$ of the $3{,}360$
annotated episodes); the agreement statistics in
Table~\ref{tab:annotation-agreement} use all $3{,}360$.

\subsubsection{Spatial perception and understanding}

\paragraph{Target perception and localization.}
This criterion records whether the executed behavior identifies the task-relevant
object and localizes it well enough to guide subsequent behavior.  Positive evidence
includes end-effector motion directed toward the correct object, camera actions
that keep the target in the region of interest, and later operations organized
around that target.  A negative judgment requires clear contrary evidence,
such as sustained motion toward an unrelated object or workspace region,
grasping a distractor, or repeatedly operating far from the target.  An isolated
exploratory motion is not sufficient to infer a localization failure.

\paragraph{Informative active exploration.}
This criterion records whether the executed behavior is consistent with actively
acquiring and using visual evidence that resolves a task-relevant ambiguity.  An
exploration opportunity is annotated when the current observation is judged
insufficient for a spatial or geometric fact needed by the next action, such as
object orientation, depth, the contact region, the relative pose of two arms,
or the visibility of a container opening.
Uncertainty about whether an executed action achieved its intended effect is
assessed under error detection rather than under this criterion.

An episode receives positive evidence only when all three pieces of behavioral
evidence are present: (i) the model selects a semantic view or camera motion in response to
the ambiguity; (ii) the resulting observation exposes information that reduces
that ambiguity; and (iii) a subsequent action or state estimate is consistent
with the new observation.  A redundant or task-irrelevant viewpoint,
or a view change whose observation does not reduce the ambiguity, is scored $0$.
Visiting an informative view and then immediately switching to an ineffective
view without using the acquired evidence is also scored $0$; a later irrelevant
view does not erase positive evidence when the informative observation was
already used.  Episodes with no assessable task-relevant ambiguity would be
marked $\bot$, but no annotated episode fell in this category: under
single-view observation some task-relevant geometric fact was always judged
underdetermined.  The criterion measures trajectory evidence of evidence
acquisition and use, not terminal task success.

\paragraph{Robot--object spatial relations (SR).}
This criterion records whether the trajectory is consistent with correct
interpretation of relative height, distance, direction, and alignment between
the gripper and the target.  Positive evidence occurs when successive translations or
rotations reduce the visible robot--object offset and it initiates contact,
grasping, or release at a geometrically plausible state.  Clear failures include
closing the gripper while it remains visibly above or beside the object,
descending at an incorrect lateral position, confusing image-plane direction
with world-axis motion, or declaring alignment while a substantial separation
remains.

\subsubsection{Robot manipulation}

\paragraph{Manipulation-semantics understanding.}
This criterion records whether the agent's behavior is consistent with correct
use of the available control commands and selection of task-relevant actions.
Positive evidence is
a purposeful use of world-axis translation, local-axis rotation, aperture, or
dual-arm commands to produce the intended physical effect.  Evidently
meaningless rotations, uncontrolled large motions, repeated open/close commands
without a contact opportunity, or actions fundamentally unrelated to the task
are inconsistent with the intended manipulation semantics.  Numerical
imprecision alone is assessed under spatial understanding rather than
automatically counted here; FG separately records pre-contact analysis.

\paragraph{Goal-directed manipulation planning.}
This criterion evaluates whether the agent constructs a coherent, goal-directed
metric action sequence from the current state to the desired interaction state.
Evidence includes an appropriate ordering of approach, alignment,
grasp/contact, lift or actuation, transport, placement, release, and retreat,
using only the phases required by the task.  Intermediate actions should
progressively reduce relevant spatial error or establish a necessary
precondition.  Large purposeless excursions, repeated movement in unrelated
regions, an approach direction incompatible with the intended contact, or a
sequence whose phases cannot lead to the goal constitute negative evidence.

\paragraph{Fine-grained pre-contact analysis (FG).}
This criterion records whether the model performs fine-grained spatial analysis
as it approaches an intended contact.  Relevant details include fingertip
placement, gripper orientation and aperture, contact location, approach or
placement height, and relative arm geometry.  Positive evidence is a
task-specific assessment of these details reflected in pre-contact decisions
and actions.  A generic instruction to manipulate carefully without
corresponding behavior is insufficient.  Every episode is scored; absence of
such analysis, including failure to reach an opportunity for it, receives $0$.
The criterion does not require successful operation completion.  A later miss,
unsuitable release, or terminal failure does not by itself negate demonstrated
pre-contact analysis, and terminal success alone does not establish it.

\subsubsection{Error recovery}

\paragraph{Error detection (ED).}
After an observable error occurs, ED is positive when subsequent behavior shows
that the agent no longer treats the preceding action as successful---for
example, it stops the current sequence, obtains another view, rechecks the
grasp, or begins a materially different adjustment.  Continuing the same plan
while ignoring a visible miss, dropped object, unstable placement, or repeated
planner rejection is negative evidence.  Error-free episodes are NA for ED.

\paragraph{Online correction (OC).}
After an observable error occurs, OC is positive only when the agent notices and
corrects a developing spatial deviation before it becomes an explicit operation
failure.  Examples include adjusting lateral alignment or gripper orientation
before closing, re-centering an object during transport, and restoring
bimanual levelness before the payload slips.  If the error is allowed to
progress into failure, or no timely corrective action is shown, OC is $0$.
Error-free episodes are NA for OC.

\paragraph{Post-failure adjustment (PF).}
After an observable error becomes an explicit failure, PF is positive when the
agent re-evaluates the new state and changes its observation strategy, geometric
hypothesis, approach, or numerical parameters before retrying.  The recovery
attempt need not ultimately complete the task: a clearly targeted, materially
revised attempt is positive behavioral evidence.  Repeating the failed action
without a meaningful change, persisting with an invalid stale plan, or
terminating despite a recoverable state is negative evidence.  If an error
occurs without a post-failure adjustment, PF is $0$; error-free episodes are NA.
OC and PF are stage-specific rather than mutually exclusive: one episode may
receive positive credit for correcting a developing deviation and later making
a separate post-failure adjustment.

\paragraph{Aggregation.}
Let
\[
E_m=\{e:g_e=1\}.
\]
For each error-recovery criterion $k\in\{\mathrm{ED},\mathrm{OC},\mathrm{PF}\}$,
the reported score is the positive rate over this shared error set,

\begin{equation}
    C_{m,k} =
    \frac{\sum_{e\in E_m} \mathbb{1}[r_{e,k}=1]}
         {|E_m|},
\end{equation}

where $|E_m|$ is the error-episode denominator used for the score.  For the
other six criteria, the denominator remains the number of episodes with a
valid opportunity to assess that criterion.  Each
first-level module is summarized by the equal-weight macro-average of its three
constituent criterion scores; these module values are descriptive summaries,
not an overall capability score.  We retain all nine criterion scores in addition to the
three module summaries so that a high value in one capability cannot conceal a
distinct failure mode in another.  We do not collapse all nine criteria
into a single scalar because their applicability sets and statistical
reliability differ.  If a capability has no applicable episode, its score is
reported as NA rather than zero and it is not silently included in a module
average.

\paragraph{Subtask progress for complex tasks.}
For a multi-stage task, terminal success is supplemented with a task-specific
checklist of observable subgoals, such as acquiring the correct object,
completing a required orientation change or handover, reaching the destination,
and establishing a stable final state.  A subtask receives credit only when its
required state transition is visible in the trajectory; merely issuing the
intended command or obtaining a successful planner return is insufficient.
Subtask completion is reported separately from terminal success and from the
nine general criterion scores.  Let $h_{m,\tau}$ and $n_{m,\tau}$ denote the
completed and scored checkpoint instances, respectively, for model $m$ on task
$\tau$.  Across the base benchmark, we use the checkpoint-level micro-average
\begin{equation}
P_m^{\mathrm{sub}}=
\frac{\sum_{\tau=1}^{14}h_{m,\tau}}
     {\sum_{\tau=1}^{14}n_{m,\tau}}.
\label{eq:subtask-completion}
\end{equation}
Thus, each checkpoint instance is equally weighted, while tasks with more
checklist stages contribute proportionally more instances.  This preserves
partial progress in long-horizon and bimanual tasks without relaxing the
benchmark's environment-defined success criterion.

\section{Agent Harness Implementation Details}
\label{app:agent-harness-details}

We evaluate every model through the same model-agnostic agent harness rather
than attaching a learned robot policy or action head.  The harness is an
independent Python implementation of a persistent tool-using agent loop,
specialized for interactive robot evaluation.  It separates model inference,
conversation state, action validation, environment execution, and run logging
so that the MLLM is the only model-dependent component of the control stack.

\begin{figure}[H]
    \centering
    \includegraphics[width=\linewidth]{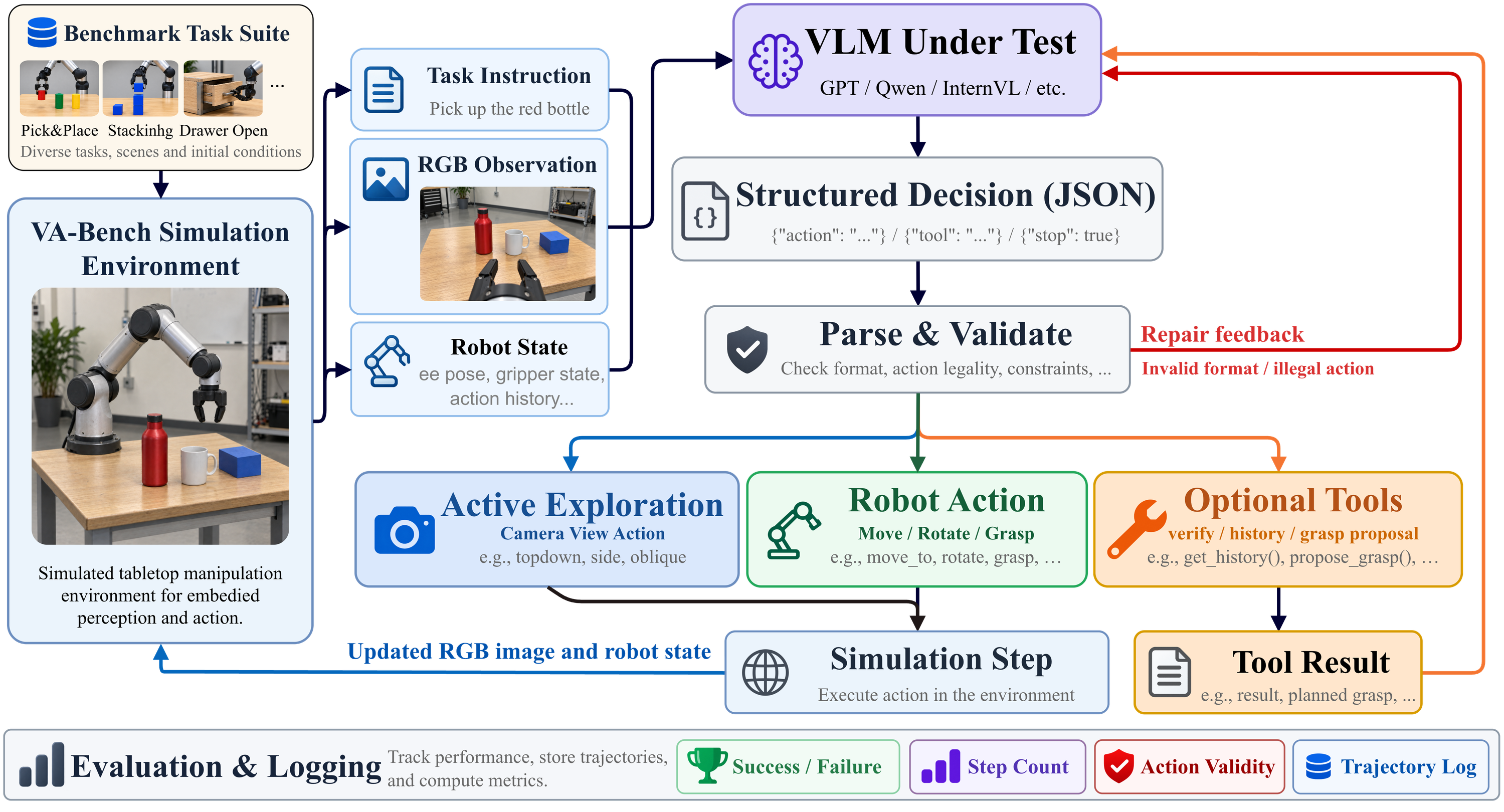}
    \caption{Schematic of the shared agent harness. Structured model decisions
    are validated and routed to camera control, robot execution, or
    non-environment tools, with observations and feedback returned to the model.
    Task, model, and command examples are illustrative. In the primary
    evaluation, robot actions are metric translations, rotations, and gripper
    open/close commands; the illustrated grasp-proposal option is not used.
    Camera actions change only the observation pose, and non-environment tools
    do not advance the simulator. Demonstration-derived summaries also
    condition the model but are omitted from the diagram.}
    \label{fig:agent-harness}
\end{figure}

\subsection{Persistent decision loop}

At each outer control decision, the harness sends the current observation
payload together with the retained conversation state and asks the MLLM for one
JSON object.  The object denotes a
non-environment tool call, a robot action, a camera action when enabled by the
observation protocol, or a stop decision.  Tool
results are returned to the same thread and may trigger another model call;
they do not advance the simulator.  Robot and camera actions pass
harness-level schema and numerical-range validation before final
environment-side arm, validity, and workspace checks.  Malformed JSON and
invalid decisions are returned to the model for a bounded number of repair
attempts, and tool-call iterations are likewise bounded.  The harness provides
adapters for OpenAI-compatible Chat Completions and Responses endpoints and
Anthropic Messages endpoints, normalizing their outputs into this common JSON
protocol rather than relying on a provider-specific robot tool interface.
The canonical active-view condition exposes camera actions.  The passive
five-view condition removes camera actions from the model-visible JSON
vocabulary because its five predefined views are supplied directly; the robot
execution stack is unchanged.

The same conversation thread is retained across the entire episode.  Both
conditions use the standard context-length limit and compaction policy; prior
robot actions, planner outcomes, action-validity feedback, and textual
messages remain in the retained history.  Each active API request contains
only the current RGB image, whereas each passive request contains only the
current fixed-order five-image bundle.  Earlier visual payloads are not
re-sent as the current observation.  Compaction leaves the append-only session
record intact while bounding the model-visible context.

For the primary base-task terminal-success comparison, the harness repeats the
fixed 20-seed suite in three independent runs for each of the 12 evaluated
models.  Run identity is retained in the session record, and aggregation is
performed only after computing the task-macro-average within each run.  The
passive five-view control, held-out transfer, composition, and trajectory
annotation analyses retain their separately identified matched single-run
protocols.

\subsection{Model observations and actions}

After reset and after every robot action, the harness acquires the next RGB
payload together with robot proprioception and retained robot/action history.
The active API sends one current image; the passive API sends only the current
control point's fixed-order five-image bundle.  An active camera action may
return a newly selected view without advancing physical state; the passive
condition has no such command.  Context limit, compaction policy, and
environment transitions are otherwise shared.
Proprioception includes gripper
state, end-effector pose, the midpoint between the physical fingers (GC), and
the current gripper-local axes expressed in the world frame.  The model also
receives the preceding action's validity and a binary trajectory-planning
outcome.  It does not receive the target object's coordinates or pose,
segmentation masks, oracle grasp points, contact state, or task waypoints.
The fixed motion-planning layer necessarily accesses simulator state required
for inverse kinematics and trajectory generation, but this privileged state is
not exposed to the MLLM.

The robot interface contains world-frame Cartesian translation increments,
gripper-local rotation increments, and open/close commands for one or two
arms.  Numerical translations and rotations are bounded; Cartesian targets
are clamped to the configured robot workspace before execution.  For a local
rotation, the adapter transforms the requested gripper-local axis into the
world frame, compensates the end-effector position so that GC remains
approximately fixed, and decomposes large rotations into planner waypoints of
at most $10^\circ$.  A fixed inverse-kinematics and trajectory layer then
attempts the requested motion.  It may reject an unreachable trajectory, but
does not select the task object, grasp, or semantic waypoint for the model.

When enabled, the camera interface combines five gripper-centered semantic
viewpoints with bounded translation, zoom, yaw, and pitch actions.  These actions directly
update the simulated camera pose: they do not model a camera arm's
reachability, collision constraints, or continuous path cost.  Camera and
robot actions both consume the episode step budget, whereas history,
self-geometry, and demonstration-retrieval tools consume inference and tool
budget without advancing the physical episode.  We therefore characterize
this interface as bounded task-conditioned viewpoint selection, not
unconstrained physical camera trajectory planning or camera-arm control.

\subsection{Demonstration and prompt conditioning}

In the demonstration-conditioned protocol, the evaluated model first inspects sampled
frames from an RGB-only video of a successful execution and writes a structured
task summary.  This track exposes no machine-readable expert action,
end-effector pose, joint state, object coordinate, contact label, or semantic
phase.  The visible
robot and object motion nevertheless conveys procedural supervision, including
operation order and approach strategy; the protocol removes direct numerical
action supervision rather than making the demonstration action-free.  A
learning gate requires the model to inspect multiple distinct frames before
its summary can be stored.  Model-specific-summary and shared-summary tracks
allow the effect of self-distilled experience to be studied separately.  The
model-specific-summary condition is the primary end-to-end protocol, whereas
the shared-summary condition is a controlled diagnostic rather than an
interchangeable input for cross-model ranking.

The model is additionally conditioned on the task instruction, generic
manipulation and safety constraints, and task-specific profiles where needed
for roles or completion conditions.  Thus, planning remains open-ended in the
chosen observation--action sequence and numerical parameters, but it is not an
unscaffolded discovery process.  No low-level ground-truth trajectory, oracle
waypoint sequence, or target-object state is included in this scaffolding.

\subsection{Execution records and outcome}

The adapter stores the clean RGB frames delivered to the model separately from
annotated debugging and recording views.  Messages and events are written to
append-only JSONL files, while image artifacts are stored separately and
referenced by the session record.  Per-step records include the selected
action, the model-provided reason, robot self-geometry, action validity, and
the planner outcome.  An action is dispatched through the adapter to RoboTwin,
which computes the next observation and invokes the task's own success check.
Consequently, reported success is the environment-defined,
physics-simulated task outcome rather than a visual judgment made by either the
MLLM or the harness.  As emphasized in Section~\ref{sec:introduction}, this is
an end-to-end system measure whose outcome also depends on the fixed planner,
action interface, and simulator dynamics.

\end{document}